\documentclass[journal]{IEEEtran}
\ifCLASSINFOpdf
\else
\fi
\usepackage{times}
\usepackage{epsfig}
\usepackage{graphicx}
\usepackage{amsmath}
\usepackage{amssymb}
\usepackage{caption}
\usepackage{subcaption}
\usepackage{capt-of}
\usepackage{upgreek}
\usepackage{footnote}
\makesavenoteenv{tabular}
\makesavenoteenv{table}
\usepackage{cuted}
\usepackage{diagbox}
\usepackage{capt-of}
\usepackage[table,xcdraw]{xcolor}
\usepackage{multicol}
\usepackage{multirow}
\usepackage[noadjust]{cite}
\usepackage{hyperref}
\hypersetup{colorlinks,allcolors=black}

\begin{document}
%
\title{Lightweight Modules for Efficient Deep Learning based Image Restoration}
%
%
%

\author{Avisek~Lahiri$^ \dagger$, Sourav~Bairagya$^ \dagger$, Sutanu~Bera, Siddhant~Haldar
        and~Prabir~Kumar~Biswas,~\IEEEmembership{Senior~Member,~IEEE}
\thanks{$^ \dagger$ First two authors share equal contribution.}
\thanks{A. Lahiri, S. Bera, S. Haldar and P.K. Biswas are affiliated to Indian Institute of Technology Kharagpur. S. Bairagya is currently affiliated to Mathworks, India. The work was done during his tenure as a M.Tech student at IIT Kharagpur. All correspondences to \href{mailto:avisek@ece.iitkgp.ac.in}{avisek@ece.iitkgp.ac.in}}
\thanks{Copyright ${^c\!/\!_o}$ 20xx IEEE. Personal use of this material is permitted. However, permission to use this material for any other purposes must be obtained from the IEEE by sending an email to pubs-permissions@ieee.org.}
}

%
%

\markboth{Journal of \LaTeX\ Class Files,~Vol.~14, No.~8, August~2015}%
{Shell \MakeLowercase{\textit{et al.}}: Bare Demo of IEEEtran.cls for IEEE Journals}
%


\maketitle
\begin{abstract}
Low level image restoration is an integral component of modern artificial intelligence (AI) driven camera pipelines. Most of these frameworks are based on deep neural networks which present a massive computational overhead on resource constrained platform like a mobile phone. In this paper, we propose several lightweight low-level modules which can be used to create a computationally low cost variant of a given baseline model. Recent works for efficient neural networks design have mainly focused on classification. However, low-level image processing falls under the \textit{`image-to-image'} translation genre which requires some additional computational modules not present in classification. This paper seeks to bridge this gap by designing generic efficient modules which can replace essential components used in contemporary deep learning based image restoration networks. We also present and analyse our results highlighting the drawbacks of applying depthwise separable convolutional kernel (a popular method for efficient classification network) for sub-pixel convolution based upsampling (a popular upsampling strategy for low-level vision applications). This shows that concepts from domain of classification cannot always be seamlessly integrated into \textit{`image-to-image'} translation tasks. We extensively validate our findings on three popular tasks of image inpainting, denoising and super-resolution. Our results show that proposed networks consistently output visually similar reconstructions compared to full capacity baselines with significant reduction of parameters, memory footprint and execution speeds on contemporary mobile devices. Codes are made available at \href{https://github.com/avisekiit/TCSVT-LightWeight-CNNs}{https://github.com/avisekiit/TCSVT-LightWeight-CNNs}\\
\end{abstract}
\begin{figure*}
    \centering
    \includegraphics[scale=0.41]{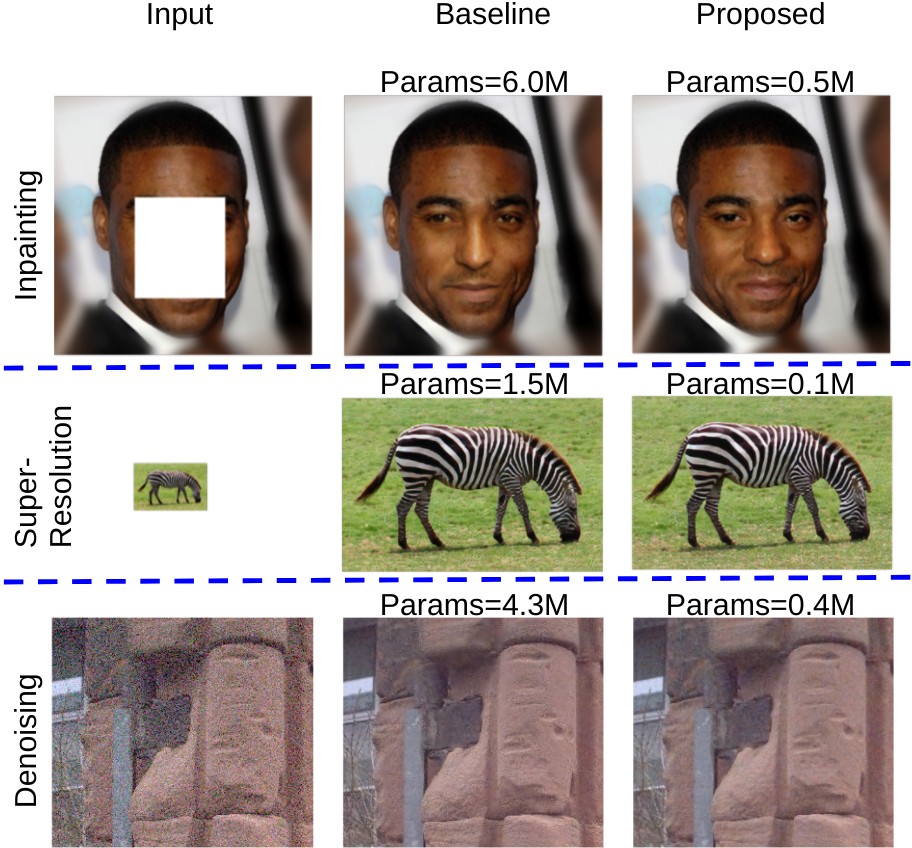}
    \caption{Visual comparison of outputs from our computationally efficient variants on three common image restoration applications. The full-scale baselines are GLCIC \cite{iizuka2017globally} for inpainting, SRGAN \cite{srgan} for super-resolution and CBDNet \cite{cbdnet} for denoising. Number on top of a figure denotes the total number of parameters (in millions) of a particular model. Best viewed zoomed in.}
\label{fig_cover}
\end{figure*}

%
\IEEEpeerreviewmaketitle
\section{Introduction}
\IEEEPARstart{I}{mage} restoration refers to recovery of clean signal from an observed noisy input.  Following the ground-breaking work of Krizhevsky \textit{et al.} \cite{alexnet} on ImageNet classification with deep neural networks, CNNs have superseded traditional methods across a variety of tasks such as object recognition \cite{resnet, szegedy2016rethinking, szegedy2015going}, detection \cite{rcnn, fastrcnn, fasterrcnn} and tracking \cite{bertinetto2016fully,he2018twofold}, action recognition \cite{carreira2017quo,hara2017learning}, segmentation \cite{maskrcnn,noh2015learning} to list a few. Image restoration frameworks also improved from the data driven hierarchical feature learning capability of deep neural networks with state-of-the-art performances on inpainting \cite{iizuka2017globally, gip, shiftnet, gfc, lahiricvpr, lahiriicip}, denoising \cite{zhang2017beyond,tian2020image,yuan2018hyperspectral}, super-resolution \cite{srgan,tai2017image}, de-hazing \cite{ren2016single}, de-occlusion \cite{zhao2017robust}, 3D surface reconstruction \cite{jackson2017large,sinha2016deep} etc. Though these deep learning based restoration frameworks yield photo-realistic outputs,
the models are computationally expensive with millions of parameters. Inference through such complex networks requires billions of floating point operations (FLOPs). This might not be seen as a problem while executing over a GPU enabled workstation; however such networks are practically not scalable to run on resource-constrained platforms such as a commodity CPU or a mobile device. However, with the proliferation of multimedia enabled mobile devices, there is an increased demand of on-device multimedia manipulations. For example, image denoising is a crucial component of imaging setup in any contemporary smartphone. Super-resolution is also an inevitable component because online  multimedia hosting sites often prefer to transmit low resolution images and videos with super-resolution performed on device so that the end user enjoys high resolution multimedia experience even on low bandwidth channel. Similarly, inpainting plays a crucial role in many downstream applications such as image editing, Augmented Reality \cite{schops2017real}, `dis-occlusion' inpainting \cite{luo2019disocclusion, lin2018recovery} for  novel view synthesis in a multi camera video capture setting \cite{ceulemans2016efficient}
to be integrated with mobile Head Mounted Displays (HMD). 
\par Executing billions of FLOPs on mobile devices leads to fast reduction of battery life with potential heating up of the device. Also, the lag encountered while executing such large models on constrained platform tends to disrupt the engagement of the user. To address the above two issues, in this paper we propose several lightweight computing units which dramatically reduce the computational cost of a given deep neural network without any visual degradation of reconstructed outputs.
\par Recently, there has been a surge of interest for designing efficient neural networks mainly for object classification and detection. However, there is a dearth of literature for efficient processing of networks concerned with low-level image restoration. Fundamentally, restoration requires the spatial resolution of input and output signal to be same and the general practice \cite{iizuka2017globally, gip, zhang2017beyond} is to follow encoder-decoder based architectures to first down-sample and later on up-sample the intermediate feature maps of the network. On contrary, classification frameworks are mainly concerned with progressive downsampling and thus efficient strategies to up-sample in a network are not discussed. Also, dense prediction tasks such as inpainting requires long range spatial information and often deploys dilated/atrous convolutions \cite{dilated_convolution} to increase the receptive field of processing. However dilated convolutions are rarely used in classification frameworks and thus recent advancements such separable convolution \cite{mobilenet} and group convolutions \cite{shufflenet} cannot be directly applied for dilated convolution operations.
\par In this paper we have mainly focused on design principles for components to be used in low-level restoration tasks. Since 3$\times$3 kernel \footnote{Ideally, it should be 3$\times$3$\times C_{in}$- where $C_{in}$ is number of input channels. For brevity of notation, henceforth, we will drop the channel dimension.} is the most commonly used kernel in contemporary low level vision applications \cite{iizuka2017globally, gip, zhang2017beyond, srgan}, we introduce \textit{`\textbf{LI}ght \textbf{S}patial \textbf{T}ransition'} layer, (\textit{LIST}), which simultaneously benefits from local feature aggregation \cite{nin} and multi-scale spatial processing \cite{szegedy2015going} and uses upto  24$\times$ fewer parameters than a similar 3$\
\times$3 convolution layer. Next, we introduce \textit{`\textbf{G}rouped \textbf{S}huffled \textbf{A}trous \textbf{T}ransition'} layer, \textit{GSAT}, which is an efficient atrous/dilated convolution layer by leveraging recent concepts of group convolution \cite{alexnet} and channel shuffling \cite{shufflenet} and each layer uses approximately 7$\times$ fewer parameters compared to an usual dilated convolution layer. While designing efficient upsampling module, we show that separable convolution kernels are inept at sub-pixel convolution \cite{shi2016real} based upsampling and we provide an analytical justification for the same. Instead we show that deterministic upsampling such as  bilinear upsampling followed by our \textit{LIST} module provides an efficient upsampling framework. Combination of these modules enable us to run image restoration models on mobiles with milli-seconds level execution speed compared to several seconds by contemporary full-scale models without any visual degradation of outputs. One of the major advantages of our proposed modules is that these can seamlessly replace commonly used computational blocks such as 3$\times$3 convolution, dilated convolution, differentiable upsampling within a given network. Thus, in this paper we refrain from proposing new end-to-end architectures; instead we select recent state-of-the-art networks and reduce the computational footprints of those networks using our lightweight layers.
In summary, our key technical contributions in this paper are:\\
-We present \textit{LIST} layer as a computationally cheaper alternative to a regular 3$\times$3 convolution layer. Each instance of \textit{LIST} can save 12$\times$ - 24$\times$ parameters. Repeated use of \textit{LIST} in a deep network leads to significant reduction of parameters and FLOPs \\
-We present \textit{GSAT} layer which implements dilated convolution on separate sparse group of channels to reduce FLOPs followed by feature mixing for enhanced representation capability. Each instance of the proposed module utilizes approximately 7$\times$ fewer parameters than a regular dilated convolution layer \\
- We present our findings on drawbacks of applying separable convolution for feature upsampling with sub-pixel convolution and provide a detailed insight for possible reason of failure. Instead, we show that deterministic upsampling followed by \textit{LIST} layer based convolution is an efficient yet accurate alternative \\
- We perform extensive study of our network components on tasks of image inpainting, denoising and super- resolution. On all tasks we achieve significant reduction in parameters and FLOPs and massive execution speed-ups on resource constrained platforms without any compromise in visual quality. Such exhaustive experiments manifest the generalizability of our processing components across a variety low-level restoration tasks.
\section{Related Works}
In recent years deep neural networks have achieved overwhelming success on a variety of computer vision tasks in which network design plays a crucial role. Executing these large models on resource constrained platforms requires efficient design strategies \cite{he2015convolutional}. Recently there has been a surge of interest in either compressing existing pre-trained big networks or designing small networks from scratch.
\subsection{Kernel Factorization}
For training small networks from scratch, factorization of kernels have been a preferred choice. The most common realization is depthwise separable convolution initially presented in \cite{sifre2014rigid} and then popularized in Inception module \cite{szegedy2015going}. Following that, it has become the backbone of many popular architectures such as MobileNet \cite{mobilenet} and MobileNet-V2 \cite{mobilenetv2}. Xception network \cite{chollet2017xception} showed how to scale up depthwise separable convolutions to outperform Inception-V3 \cite{szegedy2016rethinking}. Another popular concept of group convolution was introduced in \cite{alexnet} to distribute model parameters over multiple GPUs. Currently, it is utilized in several recent efficient networks \cite{zhang2018shufflenet, ma2018shufflenet, xie2018interleaved,sun2018igcv3}. The idea is to convert dense convolutions across all feature channels to be sparse by channel grouping and performing convolution only on grouped set of channels.
\subsection{Model Compression}
Model compression is another genre of approach for efficient inferencing by lossy compression of a pre-trained network while maintaining similar accuracy. Compression can be achieved either by pruning some of the intermediate synaptic connections in the network or by quantizing pre-trained kernels to be represented as integers or booleans. Denton \textit{et al.} \cite{denton2014exploiting} applied Singular Value Decomposition (SVD) to approximate a pre-trained network to achieve 2$\times$ inference speedup. Han \textit{et al.} \cite{han2015learning} pruned and fine-tuned a pre-trained network to identity important network connections to create a smaller network. The work was extended in \textit{Deep Compression} \cite{han2015deep} to combine network pruning with quantization. Later, \textit{`Quantized CNN'} \cite{wu2016quantized} was proposed which aimed at directly quantizing network weights during training. Chen \textit{et al.} proposed \textit{`HashedNet'}
\cite{chen2015compressing} to compress networks with hashing. 
\subsection{Task Specific Efficient Architectures}
Some recent works have focused on smarter network designs for efficient low-level vision applications. Zhang and Tao \cite{zhang2019famed} proposed a light-weight multi-scale network for single image dehazing. In \cite{effe} Ahm \textit{et al.} proposed a cascaded residual network coupled with group convolution for efficient single image super-resolution. In \cite{tan2019efficient}, Tan \textit{et al.} presented a low-cost network for unmanned aerial vehicle (UAV) noise reduction at low signal-to-noise (SNR) level. In \cite{zhang2018dcsr}, Zhang \textit{et al.} present a \textit{`mixed-convolution'} layer by merging normal and dilated convolution for image super-resolution. Kim \textit{et al.} \cite{kim2019efficient} presented dilated-Winograd transformation for a faster realization of dilated convolution. In RHNet \cite{yu2019rhnet} the authors present a dilated special pyramid pooling framework for dense object counting.
\par In this paper we mainly focus on constructing lightweight modules for training efficient networks from scratch for low-level image restoration tasks. However, the building blocks of modern efficient networks are mainly concerned with classification tasks in which essential components such as upsampling, sub-pixel convolution and dilated convolution are usually not involved. Hence, those methods are not self-sufficient for low-level computer vision applications.
\par In recent years deep learning based methods have produced phenomenal performances on a variety of low-level image restoration tasks. However majority of research has been focused on improving the visual quality without worrying much about the computational burden. In this paper we aim to realize lightweight versions of these networks which can be run on mobile devices with milli-seconds level execution time instead of multiple seconds required by full-scale baselines.
\section{Proposed Network Modules}
\subsection{`LIght Spatial Transition' layer: (\textit{LIST})}
\begin{figure}[!t]
\centering
   \begin{subfigure}{0.5\linewidth}
     \includegraphics[scale = 0.3]{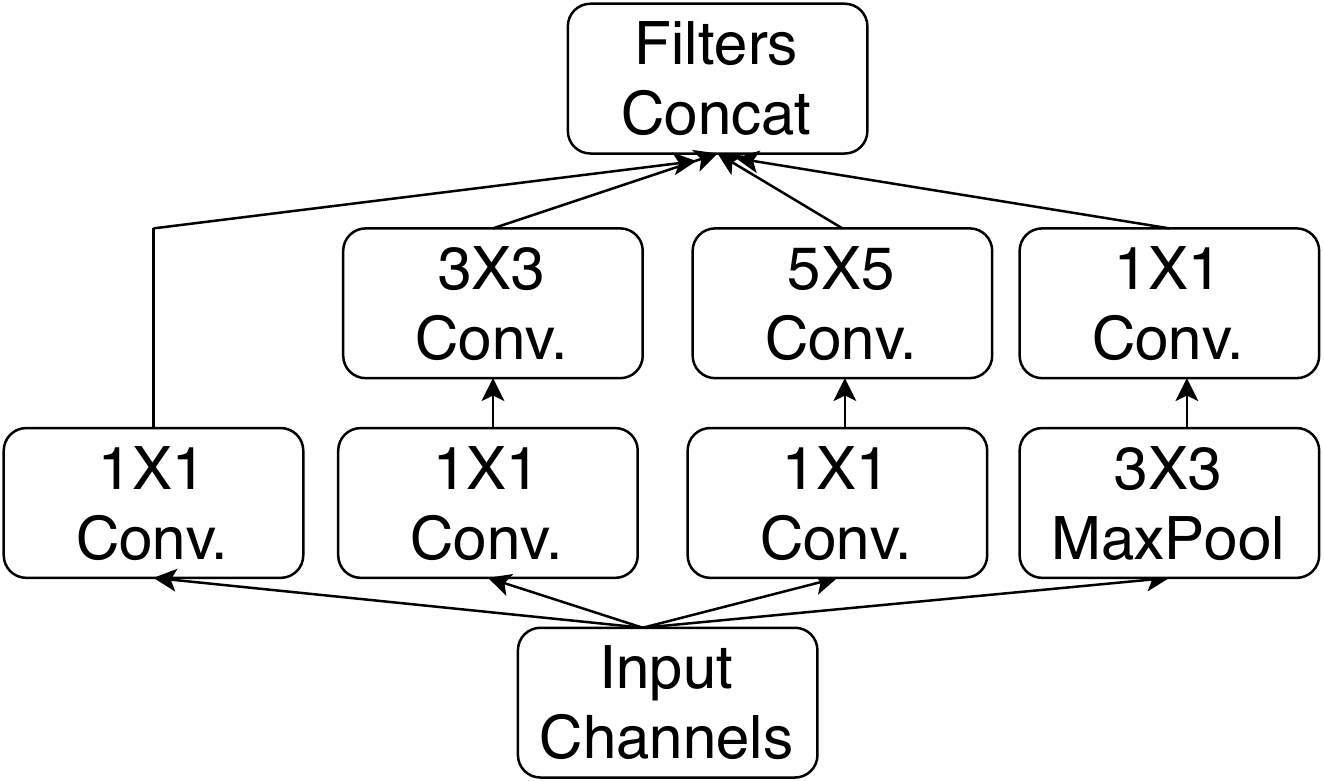}
     \caption{\scriptsize Inception module.}\label{fig_inception_module}
   \end{subfigure}\\
    \begin{subfigure}{0.55\linewidth} \centering
     \includegraphics[scale = 0.4]{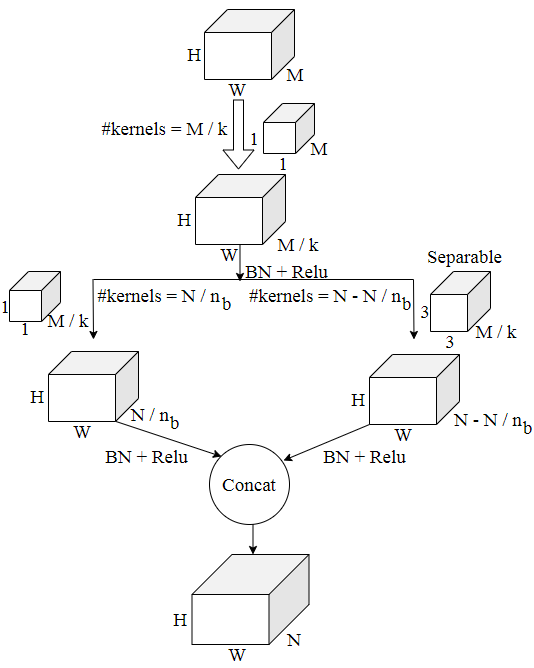}
     \caption{\scriptsize LIST module}\label{fig_list_subfigure}
    \end{subfigure}\\
   \caption{\scriptsize (a) An usual Inception module \cite{szegedy2015going}. (b) Proposed \textit{LIST}$^{M \rightarrow N}$ module as a replacement of normal 3$\times$3 kernel operating on $M$ incoming channels and yielding $N$ output channels. \textbf{BN:} Batch-normalization \cite{batchnorm} layer; \textbf{Relu(x):} $max(0,x)$.}
   \label{fig_pixel_shuffle_iccv_diag}
   \end{figure}

This section elaborates on the architectural details of \textit{LIST} layer. Pictorial representation of a \textit{LIST} layer is shown in Fig. \ref{fig_list_subfigure}. We will first discuss the driving intuitions and principles behind \textit{LIST} followed by calculating computation savings achieved by using \textit{LIST} instead of regular 3$\times$3 convolution layer. Presence of a `sub-network' capable of universal functional approximation such as multi-layer perceptron (MLP) in between two consecutive layers boosts the feature extraction capability in a CNN \cite{nin}. In \textit{LIST}, we realize this functionality by having one parallel branch of two successive layers of 1$\times$1 convolution with ReLu non-linearity in between to promote sparsity of features. Such cascades of 1$\times$1 convolution promotes parametric cross-channel pooling and enables a network to learn non-trivial transformations. \par Starting from the Inception \cite{szegedy2015going} module of GoogleNet (see Fig. \ref{fig_inception_module}), multi-path branched module has become the de facto choice for multi-scale processing of features in deep neural networks \cite{szegedy2016rethinking}. Following that, we incorporate a branch for 3$\times$3 convolution in parallel with the 1$\times$1 branch. In this case, the initial (top) 1$\times$1 layer acts an embedding layer by projecting incoming feature volume to a lower dimension and thereby reducing the FLOPs requirement for performing 3$\times$3 convolution. We further reduce the FLOPs count for 3$\times$3 convolution branch by factoring it with depthwise separable kernels. However, we deviate from the design principles of Inception by restricting the number of parallel branches inside the \textit{LIST} layer. This is motivated by the `network fragmentation' issue pointed out in \cite{ma2018shufflenet}. Parallel branches in a network creates overhead of kernel launching and synchronization resulting in reduction of execution speed. So, unlike that in Inception, we refrain from using two additional parallel branches of 5$\times$5 convolution and max-pool layer inside our \textit{LIST} layer. Apart from `network fragmentation' issue, avoiding parallel branches also benefits from reduced number of final feature channels which need to be processed by next layer- this further helps in decreasing FLOPs.\\
\subsubsection{Architecture Details}
A \textit{LIST}$^{M \rightarrow N}$ layer is meant for replacing a normal 3$\times$3$^{M \rightarrow N}$ convolution layer with $M$ input and $N$ output feature channels.
Input to a \textit{LIST}$^{M \rightarrow N}$ module is a feature volume of shape ($H$, $W$, $M$) (height, width, channels). In the first step, the input volume is pointwise convolved with $\frac{M}{k}$ number of 1$\times1$ kernels; $k$ is the compression ratio. In the second stage, these $\frac{M}{k}$ feature maps are passed to two parallel streams of 1$\times$1 and 3$\times$3 convolution. In the 1$\times$1 branch, we perform another set of pointwise 1$\times$1 convolution and output $\frac{N}{n_b}$ channels; $n_b$ is the branching factor. The 3$\times$3 branch is realized with depthwise separable kernels and  outputs $N - \frac{N}{n_b}$ channels. Outputs from 1$\times$1 and 3$\times$3 streams are concatenated (to form total $N$ channels) and passed on to the next layer.\\
   
\subsubsection{Computational Savings} ~\footnote{All throughout the paper, we consider `valid' convolution by padding zeros at the border and stride of 1 pixel; this preserves the image resolution .} \\
\textbf{Comparison to 3$\times$3 convolution:}
In this section we elaborate on the savings of parameters and FLOPs achieved by our \textit{LIST}$^{M \rightarrow N}$ layer over the usual 3$\times$3$^{M \rightarrow N}$ layer. We assume the spatial resolution of incoming and outgoing features to be H$\times$W.
Number of trainable parameters for a 3$\times$3$^{M \rightarrow N}$ is,
\begin{equation}
    P_{3\times3} = 9 M N,
\end{equation}
while the total FLOPs is,
\begin{equation}
    F_{3 \times 3} = 9 M N 
     H W~.
\end{equation}
Computations for a \textit{LIST}$^{M \rightarrow N}$ module will consist of three components- (a) 1$\times$1 convolution in Stage-1; (b) 1$\times$1 convolution in Stage-2 parallel stream; (c) separable 3$\times$3 convolution in Stage-2 parallel stream. Assuming $n_b$ = 2 (see Sec. \ref{sec_ablation}) number of parameters for a LIST$^{M \rightarrow N}$ layer is,
\begin{equation}
  P_{LIST} =   \underbrace{\frac{M^2}{k}}_\text{1X1 Stage-1}+ \underbrace{\frac{MN}{2k}}_\text{1X1 Stage-2} + ~ \underbrace{3^2 \times \frac{M}{k} + \frac{M N}{2k}}_\text{3X3 Stage-2} ~,
\end{equation}
while total FLOPs is,
$$ F_{LIST} = \underbrace{  \frac{H  W  M^2}{k}}_\text{1X1 Stage-1}
+ \underbrace{  \frac{H W MN}{2k} }_\text{1X1 Stage-2} $$
\begin{equation}
     +  \underbrace{3^2 \times   \frac{H  WM}{k} + \frac{H  W  M  N}{2k}}_\text{3X3 Stage-2}.
\end{equation}
Ratio of parameters of 3$\times$3$^{M \rightarrow N}$ to that of \textit{LIST}$^{M \rightarrow N}$ is given by,
\begin{equation}
    R_{params}^{3\times 3 | LIST} = \frac{9  N  k}{M + N + 9},
\end{equation}
\begin{equation}
\approx \frac{9  N  k}{M + N}.
\label{eq_approx_params_3x3}
\end{equation}
Since, $k$, is the compression ratio of incoming and outgoing channels to the first 1$\times$1 layer, $k >$  1. Thus, we have,
\begin{equation}
    R_{params}^{3\times 3 | LIST} > \frac{9  N}{M + N}.
    \label{eq_params_lower_bound_3x3}
\end{equation}
From Eq. \ref{eq_params_lower_bound_3x3} we get the lower bound of parameters saving by using proposed \textit{LIST} layer instead of a 3$\times$3 convolution layer. Some of the usual settings in a network are $M=N$, $M = 2N$ or $N = 2M$. After a brief hyper-parameters search (see Sec. \ref{sec_ablation}) we set $k$ = 4 and thus we achieve 18$\times$, 12$\times$ and 24$\times$ parameters saving at $M=N$, $M = 2N$ and $N = 2M$. Thus a single instance of our \textit{LIST} layer is significantly cheaper than a normal 3$\times$3 convolution layer. On a similar note, we can show that the ratio of FLOPS of 3$\times$3$^{M \rightarrow N}$ to that of LIST$^{M \rightarrow N}$ is given by,
\begin{equation}
    R_{Flops}^{3 \times 3|LIST} = \frac{9  N  k}{M + N + 9}.
\end{equation}
Since the ratio is same as what we got for parameters savings,
following the approximation done in Eq. \ref{eq_approx_params_3x3} and lower bound logic of Eq. \ref{eq_params_lower_bound_3x3}, we get similar scales of FLOPs savings as we showed for the parameters.
Stacking several layers of \textit{LIST} layer thereby helps in significant reduction of memory footprint (fewer parameters) and faster execution speed (fewer FLOPs) compared to a network realized with 3$\times$3 convolution layers.\\
\begin{figure*}[!t]
\centering
   \begin{subfigure}{0.99\linewidth}
   \centering
     \includegraphics[scale = 0.3]{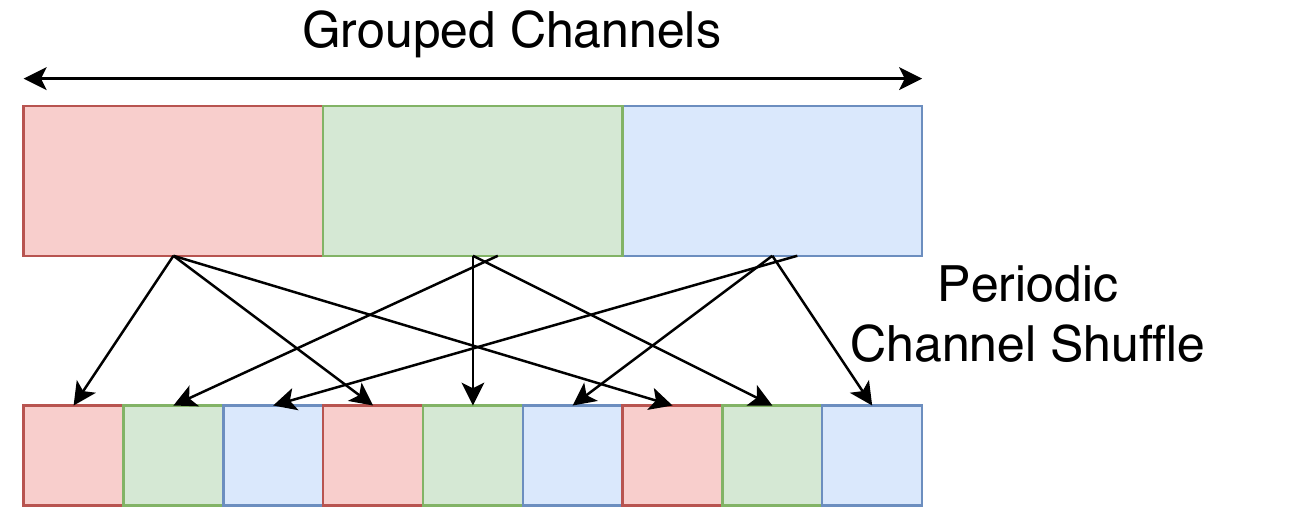}
     \caption{\scriptsize Channel-shuffle layer.}\label{fig_channelshuffle_module}
   \end{subfigure}\\
    \begin{subfigure}{0.99\linewidth} \centering
     \includegraphics[scale = 0.35]{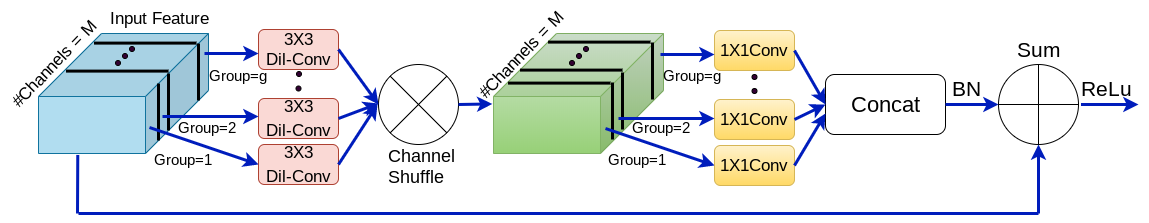}
     \caption{\scriptsize GSAT module}\label{fig_gdilated_layer}
    \end{subfigure}\\
   \caption{\scriptsize (a) Channel-shuffle layer as presented in \cite{shufflenet} (b) Proposed \textit{GSAT} layer for lightweight realization of dilated convolution. Channel Shuffle enables periodic mixing of features coming from each of the preceeding dilated convolution layers. Concat block concatenates feature volumes (along channel dimension) output from the 1$\times 1$ convolution layers. \textbf{Dil-Conv:} Dilated convolution; \textbf{BN:} Batch-normalization.}
   \label{fig_dilated_conv_layers}
   \end{figure*}
   
\textbf{Comparison to depthwise separable 3$\times$3 convolution:}
In this section we first find the condition under which proposed \textit{LIST} layer is even cheaper than the widely used depthwise separable convolution layer. We again assume 3$\times$3 convolution over a feature volume of $M$ incoming and $N$ outgoing channels and spatial resolution of H$\times$W. Number of trainable parameters for a separable 3$\times$3 convolution layer is,
\begin{equation}
    P_{3 \times 3 | sep} = 9M + MN,
\end{equation}
while total FLOPS is,
\begin{equation}
    F_{3 \times 3 | sep} = 9HWM + HWMN
\end{equation}
Ratio of parameters for a separable 3$\times$3 convolution layer to that of \textit{LIST} is,
\begin{equation}
    R_{params}^{sep -3 \times 3 |LIST} = \frac{k (N + 9)}{M + N + 9}~~\approx \frac{N k}{M + N}
    \label{eq_sep_3x3_params}
\end{equation}
If we want $R_{params}^{sep -3 \times 3 |LIST} > 1$ then we need to satisfy the following condition:
\begin{equation}
    R_{params}^{sep -3 \times 3 |LIST} > 1~~~\implies~~~~k > \frac{M}{N} + 1.
\end{equation}
So, we have the following criteria for $k$ at different ratios of $\frac{M}{N}$:
\begin{equation}
    k > \begin{cases}
    2 & if ~~M = N.\\
    3 & if ~~M = 2N.\\
    1.5 & if ~~N = 2M.
    \end{cases}
    \label{eq_cases_params_sep_conv}
\end{equation}
To satisfy all the conditions of Eq. \ref{eq_cases_params_sep_conv} we need $k>3$ which gives the lower bound of parameters savings. Since we set $k=$ 4 for all our experiments, the conditions of Eq. \ref{eq_cases_params_sep_conv} are satisfied. With $k$ = 4, from Eq. \ref{eq_sep_3x3_params} we have $R_{params}^{sep -3 \times 3 |LIST}$ = 2, 2.6 and 1.3 at $M$=$N$, $N$=$2M$ and $M$=$2N$ respectively. Similarly we can show that ratio of FLOPS of a depthwise separable 3$\times$3 layer to that of \textit{LIST} is,
\begin{equation}
    R_{Flops}^{sep-3\times3|LIST} \approx \frac{Nk}{M + N}
\end{equation}
With $M=N$, $N=2M$ or $M=2N$ we would approximately save 2$\times$, 2.6$\times$ and 1.3$\times$ FLOPs respectively.
Our \textit{LIST} layer's design has appreciably fewer parameters and FLOPs compared to even a depthwise separable realization of 3$\times$3 convolution and thus can be used as an off-the-self replacement for separable convolution layer.
\subsection{`Grouped Shuffled Atrous Transition' layer, (GSAT)}
In this section we elaborate on the design of our proposed \textit{GSAT} layer which is an efficient replacement for an usual atrous/dilated convolution layer found in numerous contemporary low-level vision applications \cite{iizuka2017globally,wei2018revisiting,hamaguchi2018effective}.
Realizing a 3$\times$3 dilated convolution is not trivially possible by our \textit{LIST} module because of the 1$\times$1 convolution in the first stage. For this we propose \textit{GSAT} layer. We mainly consider a 3$\times$3 dilated convolution with same number of incoming and outgoing channels. This is the most popular configuration in contemporary architectures. Illustration of a \textit{GSAT} layer is shown in Fig. \ref{fig_gdilated_layer}.
\par Input to the layer is a feature volume of shape H$\times$W$\times$M. Based on group convolution \cite{alexnet}, we divide the incoming $M$ channels into $g$ non-overlapping groups. Then each of the groups is individually processed by an usual dilated 3$\times$3 convolution. The initial group partitioning helps in reduction of incoming channels to individual 3$\times$3 dilated convolution layers and thereby saves on parameters and FLOPs. However, each of the $g$ groups are processed independently on a sub-group of channels without any cross-group interaction. This property weakens the representation capability of the model. Thus for cross channel interaction we perform a channel shuffling operation \cite{shufflenet} to periodically sample and stack features from each of $g$ groups. This results in an intermediate volume of shape H$\times$W$\times$M. So features from a particular group are stacked every alternate $\frac{M}{g}$ channels apart. Thus a group of $\frac{M}{g}$ channels inside the intermediate volume has features from each of the $g$ groups.
Next, to perform a cross channel interaction \cite{nin} of features we include a 1$\times$1 convolution layer. However to reduce FLOPS, we perform grouped 1$\times$1 convolution partitioned over $g$ groups. Since the channel shuffling operation already populated each of the sub-groups with features from all the 3$\times$3 dilated convolution layers, the grouped 1$\times$1 layer can now learn a non-linear transformation conditioned on all the dilated convolution layers. Thus we avoid any further channel shuffling operation. Lastly, inspired by residual connection \cite{resnet}, we add the input with the 1$\times$1 group convolution's output. To our best knowledge, this is the first realization of dilated convolution layer with grouped convolution and channel shuffling.
\subsubsection{Computational Savings}
In this section we numerically illustrate the computational benefits of using our \textit{GSAT} layer instead of usual dilated convolution layer. Number of trainable parameters for a normal 3$\times$3 dilated convolution layer is given by,
\begin{equation}
    P_{3 \times 3|dil} = 3^2 \times M^2,
\end{equation}
where $M$ is the number of incoming and outgoing channels.
For \textit{GSAT} layer, number of parameters for the first stage of grouped convolution is $\frac{3^2 M^2}{g^2} \times g=\frac{3^2 M^2}{g}$ while for the second stage of 1$\times$1 grouped convolution is $\frac{M^2}{g^2} \times g = \frac{M^2}{g}$. So, total parameters for \textit{GSAT} layer is,
\begin{equation}
    P_{GSAT} = \frac{10 \times M^2}{g}~.
\end{equation}
Ratio of parameters used in regular dilated convolution and that used by proposed \textit{GSAT} layer is,
\begin{equation}
    R_{params}^{3 \times 3| GSAT} = \frac{9g}{10}~.
    \label{eq_dilated_frac}
\end{equation}
So, we can save parameters if $R_{params}^{3 \times 3| GSAT} > $ 1, which requires $g \ge $ 2. In fact, after hyper-parameters search (see Sec. \ref{sec_ablation}) we used $g = $ 8 and thus \textit{GSAT} module requires almost 7$\times$ fewer parameters compared to normal dilated convolution layer.
\subsection{Efficient Upsampling Strategies}
Upsampling of intermediate feature maps in a network is an essential component for low-level vision tasks. However, recent frameworks for efficient network design do not discuss upsampling strategies because it is rarely required in classification frameworks. We thus devote this section for discussing possible solutions for efficient upsampling.
In recent literature transposed convolution (popular as deconvolution) \cite{deconvolution_paper} has become the de facto choice for upsampling.  However, from an image generation perspective, transposed convolution is known to render `checkboard' effects \cite{odena2016deconvolution,aitken2017checkerboard} on the final synthesized image. Thus, even though there are efforts towards making transposed convolution computationally faster \cite{szegedy2016rethinking,chollet2017xception} we explore other avenues for efficient upsampling. \\
\begin{figure}[!t]
\centering
   \begin{subfigure}{0.99\linewidth} \centering
     \includegraphics[scale = 0.25]{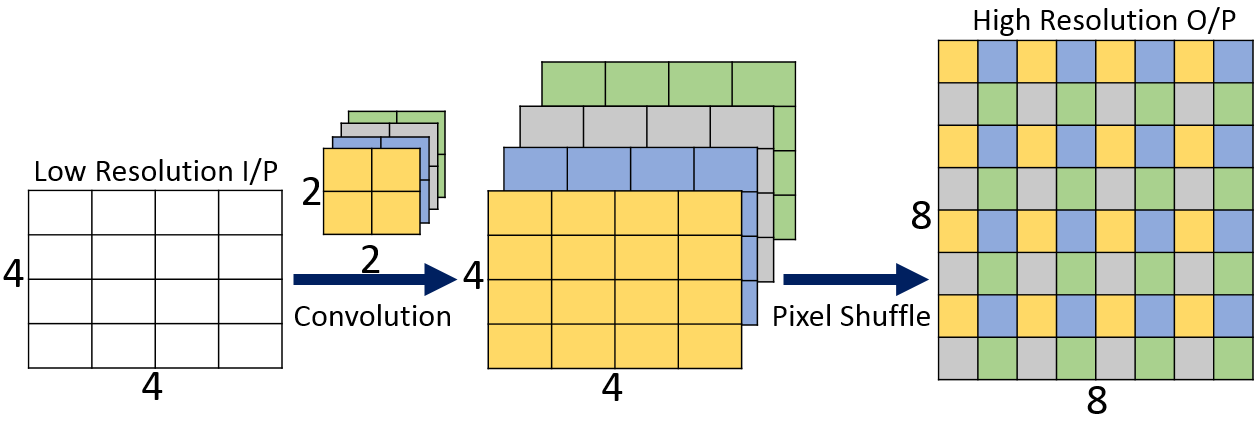}
     \caption{\scriptsize Upsampling with usual subpixel convolution followed by pixel-shuffle operation.}\label{fig_shuffle_normal_iccv}
   \end{subfigure}\\
    \begin{subfigure}{0.99\linewidth} \centering
     \includegraphics[scale = 0.25]{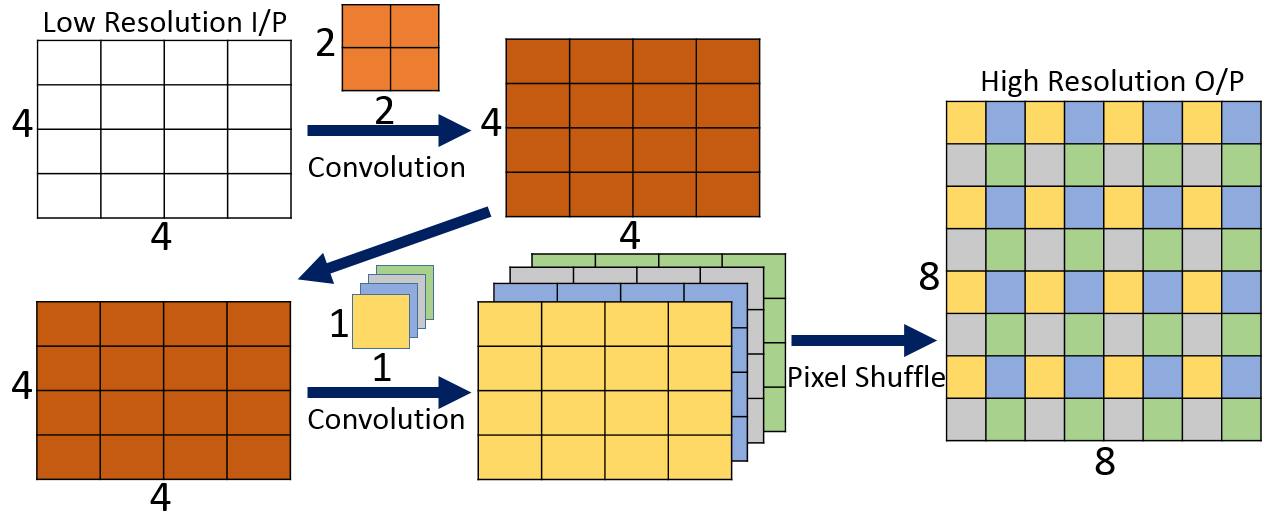}
     \caption{\scriptsize Upsampling with separable sub-pixel convolution followed by pixel-shuffle.}\label{fig_shuffle_separable_iccv}
    \end{subfigure}\\
   \caption{\scriptsize Approaches for upsampling with sub-pixel convolution and pixel-shuffle layer. Colored activation grids indicate the corresponding kernel responsible for activating that grid point on feature map.}
   \label{fig_pixel_shuffle_iccv_diag}
   \end{figure}
\subsubsection{Failure of Separable Kernels for Sub-Pixel Convolution}
\label{sec_subpixel_with_caution_supple}
Sub-pixel convolution based upsampling is a preferred paradigm of upsampling specifically for image generation tasks because of its demonstrated ability to get rid of `checkboard' artifacts introduced by transposed convolution layer.
In this section we elaborate on our initial failed attempt of applying (see Fig. \ref{fig_subpixel_fail} for failed inpainting results) separable kernels for sub-pixel convolution based upsampling and provide justifications for the same.  
\par It can be shown that, for an upscale factor of, $r$, a sub-pixel convolution with kernel shape $(k, k, o\times r^2, i)$ ( height, width, \# of output channels, \# of input channels) is equivalent to a that of a transposed convolution by a kernel of shape $(k, k, o, i)$.  After sub-pixel convolution, the $o\times r^2$ channel elements are periodically shuffled to upscale feature maps by factor of $r$ along height and width. See Fig. \ref{fig_shuffle_normal_iccv} for visualization. Refer to  \cite{shi2016real, shi2016deconvolution} for more detailed derivation.
\par  From the theory of sub-pixel convolution we know that with an upscale factor of 2, sub-pixel convolution can learn to represent $n$ feature maps in LR (low resolution space) which are equivalent to $\frac{n}{4}$ feature maps in HR (high resolution space). We will show that this essentially means both networks have same run time complexity but sub-pixel convolution has more parameters. Let us consider a general case where shape of input volume at layer $l-1$ is (height, width, depth) =  $(\frac{H}{2}, \frac{W}{2}, c_{l-1})$. The target is to upscale this to spatial resolution of $H \times W$, for next layer, $l$. Let, for sub-pixel convolution we choose kernels of shape $(k, k, c_l, c_{l-1})$. Then for counterpart of HR model (which first does deterministic up-scaling followed by convolution in HR space itself), kernel sizes will be $(k, k, \frac{c_l}{4}, c_{l-1})$. Total FLOPs for sub-pixel convolution is,
 \begin{equation}
     F_{LR} = k^2 \times \frac{H}{2} \times \frac{W}{2} \times c_{l-1} \times c_l ~.
 \end{equation}
  The number of trainable parameters for LR model is,
  \begin{equation}
      |\theta|_{LR} = k^2 \times c_{l-1} \times c_l ~.
  \end{equation}
   For convolution in HR, total FLOPs,
   \begin{equation}
       F_{HR} = k^2 \times H \times W \times c_{l-1} \times \frac{c_l}{4} ~,
   \end{equation}
    and number of parameters, 
    \begin{equation}
        |\theta|_{HR} = k^2 \times c_{l-1} \times \frac{c_l}{4}~. 
    \end{equation}
    So, important observation is that FLOPs for both LR and HR models are same but LR model has more parameters and thus greater representation capability.
\par Let us now examine what will happen when we try to realize separable sub-pixel convolution. See Fig. \ref{fig_shuffle_separable_iccv} for a visualization. In the first stage, we need kernels of shape $(k, k, c_{l-1}, 1)$ (height, width, output channels, input channels).  In this stage, total FLOPs,
\begin{equation}
F_{LR|sep_1} = k^2 \times \frac{H}{2} \times \frac{W}{2} \times c_{l-1}~,
\end{equation}
 and number of parameters,
 \begin{equation}
     |\theta|_{LR|sep_1} = k^2 \times c_{l-1}~.
 \end{equation}
  In the next stage we need kernels of shape $(1, 1, c_l,  c_{l-1})$. Total FLOPs in this stage,
  \begin{equation}
      F_{LR|sep_2} = c_{l-1} \times \frac{H}{2} \times \frac{W}{2} \times c_l~,
  \end{equation}
   and number of trainable parameters,
   \begin{equation}
       |\theta|_{LR|sep_2} = c_{l-1} \times c_l~.
   \end{equation}
    So, total FLOPs for separable LR model, $F_{LR|sep} =F_{LR|sep_1} + F_{LR|sep_2}$ and total number of parameters, $|\theta|_{LR|sep} = |\theta|_{LR|sep_1} + |\theta|_{LR|sep_2}$. Now consider the ratio,
\begin{equation}
\frac{|\theta|_{LR|sep}}{|\theta|_{HR}} = 4 \left [\frac{1}{k^2} + \frac{1}{c_l} \right ]
\end{equation}
$\frac{|\theta|_{LR|sep}}{|\theta|_{HR}} < 1$ always and thus we see that converting a sub-pixel convolution to a separable paradigm reduces its representation prowess with respect to a convolution in HR. Similarly, if we compare the FLOPs by $\frac{F_{LR|sep}}{F_{HR}} = 4 \left [\frac{1}{k^2} + \frac{1}{c_l} \right ]$, separable sub-pixel convolution is computationally cheaper. But because of its reduced representation capability, it is not recommended for practical applications.\\
\subsubsection{Deterministic Upsampling + Convolution}
\label{sec_separable_resize_conv}
One way to mitigate `checkboard' effect is to disentangle upsampling and convolution operations \cite{odena2016deconvolution}. An usual procedure is to use some deterministic upscaling followed by convolution in the high resolution space. This has worked well in applications such as super resolution \cite{dong2014learning} and inpainting \cite{gip}. But, when implemented in naive version, this increases the computational cost. For example, if we do a bilinear upscaling by 4$\times$ followed by convolution, there is a quadratic increase of feature size but `same information content' (if we count the number of floats). This makes bilinear upsampling + convolution almost 4$\times$ costlier than transposed convolution.
We optimize this concept by first upsampling with bilinear interpolation followed by an efficient convolution block realized by the proposed \textit{LIST} layer. This is our preferred method for efficient upsampling.
\subsection{Downsampling in Network}
To maintain the homogeneity in network design we prefer to realize spatial downsampling with \textit{LIST} layer. However, strided convolution is not trivially possible by \textit{LIST} module because of initial 1$\times$1 convolution stream. So, we first down-sample feature maps with bilinear interpolation and follow up with \textit{LIST} based efficient convolution. 
\section{Experiments and Results}
We organize our results as follows. In Sec. \ref{sec_inpainting}, we initially  perform extensive studies to select the hyper parameters governing the design choices for various proposed modules based on image inpainting. We systematically investigate the role of individual components towards reduction of parameters and FLOPs. This is followed by comparison with recent full capacity inpainting baselines and compressed models realized with MobileNet \cite{mobilenet}, ShuffleNet \cite{shufflenet} and ShuffleNetV2 \cite{ma2018shufflenet}.
\par Next, with our understanding of best network configurations we compare applicability of our proposed layers on image denoising (Sec. \ref{sec_denoising}) and image super-resolution (Sec. \ref{sec_sr}). It is encouraging to note that the proposed layers are quite insensitive to hyper parameters across different tasks which allows us to reuse the same set of hyper parameters across all the three above mentioned applications without degradation of visual quality.
\subsection{Image Inpainting}
\label{sec_inpainting}
We select the globally and locally consistent image inpainting model, GLCIC \cite{iizuka2017globally} as our baseline for image inpainting. Currently, GLCIC serves as a strong Generative Adversarial Network (GAN) \cite{goodfellow2014generative} based contemporary baseline for inpainting and we aim at realizing a lightweight version of GLCIC using our proposed layers. A GAN framework consists of two deep neural nets, generator, $G_{\theta_G}$, and discriminator, $D_{\theta_D}$. The task of the generator is to generate an image, $x\in \mathcal{R}^{H\times W \times 3}$ with a latent noise prior vector, $z\in \mathcal{R}^d$, as input. $z$ is sampled from a known distribution, $p_z(z)$. A common choice \cite{goodfellow2014generative} is, $z\sim \mathcal{U}[-1,1]^d$. The discriminator has to distinguish real samples (sampled from $p_{data}$) from generated samples. Discriminator and generator play the following two-player min-max game on $V(D_{\theta_D},G_{\theta_G})$: 
$$\underset{G_{\theta_G}}{min}~~ \underset{D_{\theta_D}}{max}~~ V(D_{\theta_D}, G_{\theta_G}) = \mathbb{E}_{x\sim p_{data}(x)}[\log D_{\theta_D}(x)]$$
\vspace{-3mm}
\begin{equation}
+\mathbb{E}_{z\sim p_{z}(z)}[1 - D_{\theta_D}(G_{\theta_G}(z))].
\label{eq_gan_main_goodfellow}
\end{equation}
At the core, GLCIC comprises of repeated applications of 3$\times$3 convolution, 3$\times$3 dilated convolution and transposed convolution layers. Please refer to \cite{iizuka2017globally} for details of the architecture. We replaced the corresponding layers with proposed \textit{LIST}, \textit{GSAT} and \textit{LIST} based upsampling layers.\\
\textbf{Automated Visual Quality Metric:} Manually analyzing the perceptual quality of reconstruction by different models is not feasible. Recent works \cite{srgan, gip} have shown that PSNR and MS-SSIM metrics are not suitable for evaluating quality of adversarial loss guided reconstructions. Analyzing the quality and diversity of GAN samples is still an open research topic. Recently Fr\'echet Inception Distance (FID) \cite{heusel2017gans} was proposed for quantifying quality and diversity of GAN samples. Lower FID value indicates overall better quality and diversity of generated samples. For automated screening of models, we use FID as the base metric.\\
\textbf{Datasets:} We experimented on CelebA (128$\times$128) \cite{celeba}, CelebA-HQ (256$\times$256) \cite{karras2017progressive}, Places2 (256$\times$256) \cite{zhou2018places} and DTD (256$\times$256) \cite{dtd}. For CelebA, hole sizes greater than 48$\times$48 occludes almost entire face and thus maximum training hole size is 48$\times$48 at random location. For comparing FID during evaluation, a randomly positioned hole (but same for all models for a given image) of 48$\times$48 is considered. At 256$\times$256 image resolution, the maximum hole size of 96$\times$96 is considered during training and FID is reported at hole size of 96$\times$96. From CelebA, CelebA-HQ, Places2, and DTD we kept 20000, 10000, 20000, and 1000 (converted to 4000 with horizontal and vertical flip) samples for testing.\\
\textbf{Training Details:} In practice, we follow the  stagewise training procedure as presented in \cite{iizuka2017globally}. In Stage-1, we pre-train the inpainting (generator) network alone with $MSE$ (Mean Squared Error) loss for $T_1$ iterations. In Stage-2, we freeze the parameters of inpainting network and pre-train the critic (discriminator) network to distinguish between real and inpainted samples for $T_2$ iterations using cross-entropy loss. In Stage-3, both completion and critic networks are iteratively updated under the min-max GAN game formulation \cite{goodfellow2014generative} for $T_3$ iterations.\\ 
\textbf{Implementation Details}
We first discuss how we select design hyper parameters of our network modules such as \textit{LIST} and \textit{GSAT}. For a given parameter setting, we train on CelebA dataset and evaluate the FID on CelebA validation set (10000 samples). Due to lack of massive computational resources, we run parameter search sweep only on CelebA and adopted our understanding on other datasets. It is encouraging to see that lessons learned from CelebA generalize well to other datasets also.  We set $T_1$, $T_2$ and $T_3$ to 10$^6$, 10$^5$ and 10$^6$ iterations. Mini batch gradient descent based optimization is performed with ADAM \cite{adam} optimizer with batch size = 64. Following \cite{srgan}, we perform paired two-sided Wilcoxon signed-rank tests and significance level set to 10$^{-4}$.\\
\begin{table}[!t]
\centering
\scriptsize
\caption{\scriptsize FID scores on CelebA validation set by networks controlled by different settings of bottleneck ratio, $k$ and branching factor, $n_b$ of proposed \textit{LIST} module.}
\label{table_select_k_nb}
\begin{tabular}{l|llllllll}\hline\hline
 & \multicolumn{8}{c}{$\frac{1}{k}$ = 0.25}  \\\hline
$\frac{1}{n_b}$  & 0.1 & 0.2 & 0.3 & 0.4 & 0.5 & 0.6 & 0.7 & 0.8 \\\\
FID & 6.93 & 6.95 & 6.98 & 7.10 & 7.11 & 8.92 & 10.23 & 14.25 \\\hline
 & \multicolumn{8}{c}{$\frac{1}{n_b}$ = 0.5} \\\hline
$\frac{1}{k}$ & 0.1 & 0.2 & 0.3 & 0.4 & 0.5 & 0.6 & 0.7 & 0.8 \\\\
FID & 8.11 & 7.31 & 6.92 & 6.85 & 6.83 & 6.80 & 6.78 & 6.76\\\hline
\end{tabular}
\end{table}
\begin{table}[!t]
\centering 
\scriptsize
\caption{\scriptsize FID scores on CelebA validation set with different variants of our models controlled by number of groups ($g$) in the proposed \textit{GSAT} module.}
\label{table_fid_group_dilation}
\begin{tabular}{cccccc}\hline\hline
$g=1$ & $g=2$ & $g=4$ & $g=8$& $g=16$& $g=32$ \\\hline
6.72 & 6.75 & 6.80 & 7.09 & 10.23 & 20.31\\\hline
\end{tabular}
\end{table}
\begin{figure}[!t]
\centering
\includegraphics[scale=0.4]{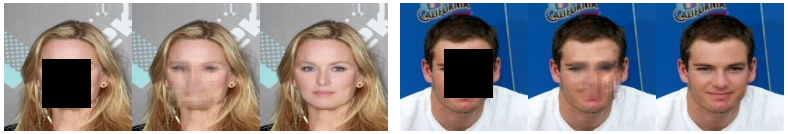}
\caption{\scriptsize Failure of separable sub-pixel convolution based upsampling (proposed model, $M_2$). For each triad, Left: Masked Image, Middle: Output with model $M_2$ guided decoder. Right: Output with decoder having regular sub-pixel convolution (proposed model $M_1$) based upsampling. Best viewed zoomed in.}
\label{fig_subpixel_fail} 
\end{figure}
\subsubsection{Hyper-parameters Search}~\\
\label{sec_ablation}
\textbf{Design parameters for \textit{LIST} module:} A \textit{LIST} module is characterized by the two hyper parameters, $k$ and $n_b$. Firstly, we study the effect of reducing 3$\times$3 kernels in the network by varying $\frac{1}{n_b}\in\{0.1, 0.2, 0.3, 0.4, 0.5, 0.6, 0.7\}$. To keep things constant, dilation layer for each case was realized with normal dilated convolution and $
\frac{1}{k}$ fixed at 0.25. In Table \ref{table_select_k_nb} we report FID metrics on CelebA validation set at a hole size of 48$\times$48. Decreasing $\frac{1}{n_b}$ (pushing more computations to 3$\times$3 stream) less than 0.5 does not improve FID appreciably but increases the model parameters while FID deteriorates briskly with increase of $\frac{1}{n_b}$ (pushing more computations to 1$\times$1 stream). We thus keep $n_b$ = 2 in our further experiments. Such a balance of channels along two parallel processing streams is also recommended in \cite{pinheiro2016learning, luc2016semantic}. Next, we sweep over different settings of $\frac{1}{k}$ at a fixed $\frac{1}{n_b}$ = 0.5. Increasing $\frac{1}{k}$ improves the representation efficacy of the Stage-1 1$\times$1 layer and thus aids in FID improvement but at a cost of higher parameters. With $\frac{1}{k} \ge$ 0.35, FID improvement almost saturates. 
\par Finally, to find a suitable threshold of FID aligned with human perception, we showed 100 inpainted images of five models with FID $\in [6.5, 7.5]$ (model with FID $\ge$ 8 are perceptually not acceptable) to five independent raters who were asked to rate a given image in $[1,5]$; 5: excellent and 1: bad quality. The difference of mean scores of models with FID $\le$ 7.0 were statistically insignificant. With $\frac{1}{n_b}$ = 0.5, from Table \ref{table_select_k_nb} we see that $\frac{1}{k}$ = 0.30 yields model in the regime of FID $\approx$ 7.0. Since channel counts in deep nets are usually of the form of $2^r$, $r \in \mathbb{N}$, we proceed in the remaining paper with $k$ = 4 ($\frac{1}{k}$ = 0.25) and $n_b$ = 2 ($\frac{1}{n_b}$ = 0.5).\\\\
\textbf{Number of Groups in \textit{GSAT} layer:} Our proposed \textit{GSAT} module is characterized by number of groups, $g$, for the group convolution layers. For simplicity of parameter sweep, we keep the group numbers same for dilated 3$\times$3 and 1$\times$1 stages. In Table \ref{table_fid_group_dilation} we report FID scores on CelebA validation set for different values of $g$. For other layers, all models used \textit{LIST} with $k$ = 4 and $n_b$ = 2 as discussed in previous section. A smaller value of $g$ indicates more computational load on the initial 3$\times$3 layers and subsequent better FIDs. However, at $g$ = 8, we get FID $\approx$ 7.0, which is perceptually acceptable. On the contrary, increasing $g$ creates many independent feature volumes and the combined channel shuffle and 1$\times$1 group convolution is not able to properly amalgamate the groups leading to higher FID. So, for future experiments we set $g$ = 8 for \textit{GSAT} layers.\\
\begin{table}[!t]
\centering
\scriptsize
\caption{\scriptsize Different variants of proposed light-weight inpainting models. Variations of models are achieved by different strategies to realize 3$\times$3 convolution layers, dilated/atrous convolution layers and upsampling in decoder sections. \textit{LIST}: 3$\times$3 convolution realized with proposed \textit{LIST} layer; \textit{GSAT}: Proposed Grouped-Shuffled convolution based dilated convolution instead of normal dilated convolution; DS: Depthwise separable 3$\times$3 convolution; BiL: Bilinear upsampling.}
\label{table_models_definitions}
\begin{tabular}{lllccc}\hline\hline
 Model& 3X3 & Upsampling & Dilation & Params (10$^6$) & FLOPs (10$^9$)\\\hline
$M_1$ & DS & \begin{tabular}[c]{@{}l@{}}Pixel Shuffle\\ (Normal Conv.)\end{tabular} & Normal & 3.42 & 33.1 \\
$M_2$ & DS & \begin{tabular}[c]{@{}l@{}}Pixel Shuffle\\ (Separable  Conv.)\end{tabular} & Normal & 2.93 & 27.1 \\
$M_3$ & DS & BiL + DS & Normal & 2.81 & 26.9 \\
$M_4$ & \textit{LIST} & BiL + DS & Normal & 2.63 & 24.8\\
$M_5$ & \textit{LIST} & BiL + \textit{LIST} & Normal & 2.61 & 24.0\\
$M_6$ & \begin{tabular}[c]{@{}l@{}}\textit{LIST}\end{tabular} & BiL + \textit{LIST} & \begin{tabular}[c]{@{}l@{}}\textit{GSAT}\end{tabular} & \textbf{0.54} & \textbf{7.4}\\\hline
\end{tabular}
\end{table}
\subsubsection{Different Speedup Variants}
In Table \ref{table_models_definitions} we define the proposed architecture variants and compare the associated parameters and FLOPs. Such analysis gives a foundation to appreciate the effect of a given speedup technique.
In Table \ref{table_fid_compare_full_capacity} we compare the FID scores of different proposed models with full-scale baseline models. Some of the key lessons from Tables \ref{table_models_definitions} and \ref{table_fid_compare_full_capacity}:\\
-- Comparing $M_3$ and $M_4$: Proposed \textit{LIST} layer is a much more efficient alternative to depthwise separable 3$\times$3 convolution layer, but both models have similar reconstruction performances.\\
-- Comparing $M_5$ and $M_6$: Proposed \textit{GSAT} layer used in $M_6$ as an alternative for normal dilated convolution layer significantly helps in reduction of parameters without hampering visual quality. Since, $M_6$ combines both \textit{LIST} and \textit{GSAT} layers, it is our preferred proposed model unless otherwise stated.\\
-- As per our theoretical justification, a network with separable sub-pixel convolution (model $M_2$) performs worse than a network with normal convolution based sub-pixel convolution (model $M_1$) as reflected by higher FID scores of $M_2$. Also, see Fig. \ref{fig_subpixel_fail} for visualizing such failures.\\
-- Comparing $M_1$ and $M_3$: Model, $M_3$, with bilinear upsampling + separable convolution has fewer FLOPs than $M_1$ (upsampled with sub-pixel convolution) while having similar FID. Thus it is prudent to have efficient bilinear upsampling which we improve further with proposed $LIST$ based upsampling in $M_5$.\\
\subsubsection{Comparing with full-scale baselines}
\begin{figure*}
    \centering
    \includegraphics[scale=0.3]{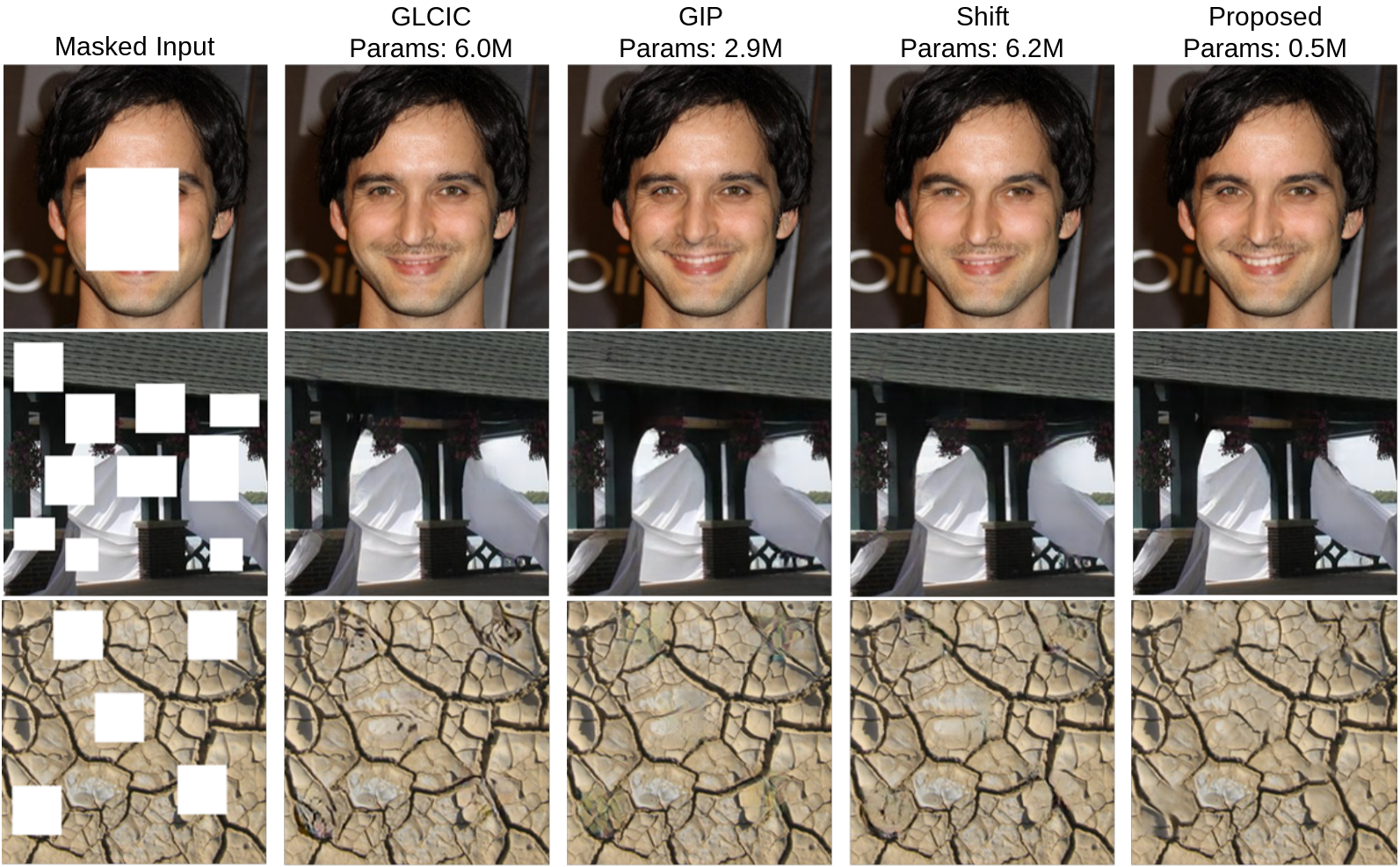}
    \caption{\scriptsize Visual comparison on inpainting on CelebA-HQ (Row 1), Places2 (Row 2) and DTD (Row 3). GLCIC \cite{iizuka2017globally}, GIP \cite{gip} and Shift \cite{shiftnet} are full-scale baselines. Our proposed model is significantly cheaper in terms on parameters yet generates similar quality reconstructions. Best viewed when zoomed in.}
    \label{fig_combined_inpaint}
\end{figure*}

Since we design all our smaller models based on the architecture of GLCIC \cite{iizuka2017globally}, it is fair to compare performances only with GLCIC as baseline. However,  for initial benchmarking of our model designs we also compared against recent state-of-the-art deep learning based models of
 GIP \cite{gip} and Shift \cite{shiftnet}. \\
\textbf{Reduction in Computation}
In Table \ref{table_compare_baselines_inpaint} we report the parameters count, FLOPs and mobile memory size. Our preferred model, $M_6$ achieves almost 91\% ( = $\frac{6.2-0.54}{6.2} \times 100 \% $) relative parameters savings compared to the parent framework of GLCIC with 88.6\% and 93.5\% relative savings in FLOPs.\\
\textbf{Comparison of Reconstruction} 
 In Table \ref{table_fid_compare_full_capacity} we report FID metrics of the comparing methods @256$\times$256 on CelebA-HQ, Places2, and DTD datasets. We did not find any significant difference of FID between any of our models (except $M_2$) and the full-scale baselines. In Fig. \ref{fig_combined_inpaint} we provide some inpainting examples by GLCIC, GIP, Shift and our preferred proposed model, $M_6$. Clearly, the reconstruction qualities of our proposed smaller model are indistinguishable from full-scale baselines. \\
\begin{table}[!t]
\centering
\scriptsize
\caption{\scriptsize Comparing FLOPs, number of trainable parameters and model sizes of inpainting models. Full capacity baseline models of GLCIC \cite{iizuka2017globally}, GIP \cite{gip} and Shift \cite{shiftnet} are compared against variants of our proposed efficient models, $M_1$, $M_2$, $M_3$, and $M_6$ derived from GLCIC.}
\label{table_compare_baselines_inpaint}
\begin{tabular}{llll|llll}\hline\hline 
&GLCIC & GIP & Shift & $M_1$ &$M_2$ &$M_3$ & $M_6$ \\\hline
FLOPs (10$^9$) & 65.0 & 41.2 & 70.1 &33.1& 27.1 & 26.9 & \textbf{7.4} \\
Params (10$^6$) & 6.02 & 2.98 & 6.24 &3.42 & 2.93 &2.81 & \textbf{0.54}  \\\hline
\end{tabular}  
\end{table}
\begin{table}[!t]
\centering
\scriptsize
\caption{\scriptsize FID (lower is better) of full-scale baselines of GIP \cite{gip}, Shift\cite{shiftnet} and GLCIC \cite{iizuka2017globally} and proposed efficient variants, $M_1$-$M_6$ derived from GLCIC.}
\begin{tabular}{l|lllllll}\hline\hline
Dataset & \multicolumn{6}{c}{Full-Scale Baselines} \\\hline
 &&& GIP & Shift &GLCIC &\\\hline
CelebA &&&  6.98 & 6.95   & 7.00 &\\
CelebA-HQ &&&   8.12 & 8.00 & 8.05 &  \\
Places2 &&&  13.10 & 13.00 & 13.25  &\\
DTD &&&    6.00 & 6.01 & 6.04  &\\
\hline
 & \multicolumn{6}{c}{Proposed Efficient Variants}   \\\hline
 & $M_1$ & $M_2$ & $M_3$ & $M_4$ & $M_5$ & $M_6$\\
CelebA & 7.12 & 23.41 & 7.11 & 7.09 & 7.11  & 7.03   \\
CelebA-HQ & 8.09 & 27.21 & 8.16 & 8.14 & 8.10 & 8.09    \\
Places2 & 13.27 & 30.41 & 13.27 & 13.29 & 13.30 & 13.39   \\
DTD &6.04 & 18.21 & 6.84 & 6.06 & 6.06 & 6.04   \\
\hline
\end{tabular}
\label{table_fid_compare_full_capacity}
\end{table}
\begin{table}[!t]
\centering
\scriptsize
\caption{\scriptsize Mean Opinion Score (MOS) by different full-scale baselines of GIP \cite{gip}, Shift \cite{shiftnet}, GLCIC \cite{iizuka2017globally} and our  cheaper variants, $M_3$ and $M_6$ derived from GLCIC. Last column shows MOS on original images.}
\label{table_mos}
\begin{tabular}{llllll|l}\hline\hline
Dataset & GIP & Shift & GLCIC & $M_3$& $M_6$ & Original \\\hline
CelebA  & 4.24 & 4.30 &4.25 &4.18 &4.22 & \textbf{4.42} \\
CelebA-HQ  & 4.17 & 4.20 &4.14 &4.13& 4.18 & \textbf{4.72} \\
Places2  & 4.00 & 3.97 &3.95 & 3.93 &3.98 & \textbf{4.60} \\
DTD  & 4.36 &4.38  & 4.41 & 4.32&4.40 & \textbf{4.55}\\\hline 
\end{tabular}  \vspace{-3mm}
\end{table}
\textbf{Mean Opinion Score Testing (MOS):} To further bolster our findings, we conducted MOS testing to visually quantify the quality of inpainting by different models. Raters were asked to rate an inpainted image in the scale of 1 (bad quality) to 5 (excellent quality). Total of 20 raters were selected for the study. From each dataset, each rater was shown 50 inpainted images by  GIP, Shift, GLCIC and proposed models $M_3$, $M_6$ models. Original images were also rated. So, each rater rated 1200 samples (4 datasets $\times$ 6 models $\times$ 50 images). We used two random positioned holes (but same across all model for an image) of 64$\times$64. In Table \ref{table_mos} we report the MOS for each dataset. Encouragingly MOS also follows the trend of FID scores. Similar to our FID findings, the difference of MOS scores between our models and any of the full-scale baselines are not significant.\\
\begin{table}[!t]
\centering
\scriptsize
\caption{\scriptsize Comparing average run time (in seconds) on different mobile devices (first three rows) and a commodity CPU (last row). Full-scale baseline models of GLCIC \cite{iizuka2017globally}, GIP \cite{gip} and Shift \cite{shiftnet} are compared against our proposed efficient variants $M_1$, $M_2$, $M_3$, and $M_6$ derived from GLCIC.}
\label{table_runtime_cpu_mobile_supple}
\begin{tabular}{llll|llll}\hline\hline 
Device&GLCIC & GIP & Shift & $M_1$ &$M_2$ &$M_3$ & $M_6$ \\\hline
Mi A1  & 8.2 & 5.5 & 9.2 &1.9&1.6 &1.4 & \textbf{0.8} \\ 
Motorola & 8.0 & 5.4 & 9.1 &1.7&1.5 &1.2 & \textbf{0.7} \\
Asus  & 5.8 & 3.1 & 6.2 & 1.0 & 0.8 &0.6 & \textbf{0.35} \\\hline
CPU & 2.1 & 0.8 & 1.4 & 0.49 & 0.42 & 0.38 & \textbf{0.30}\\ 
\hline
\end{tabular}  
\end{table}
\subsubsection{Execution Time on Mobile and CPU}
For comparison on mobile we select two low-end mobile device namely, Mi A1 and Motorola G5 S-Plus and one high-end Asus Zenfone 5Z all running on Android operating system. Mi and Motorola has 1.9GHz Qualcomm  Snapdragon 625 processor while Asus has 2.8 GHz snapdragon 845 processor. TensorFlow Lite \cite{tflite} was used for mobile execution and the framework was executed on a single thread. In Table \ref{table_runtime_cpu_mobile_supple} we report the execution times on 256$\times$256 resolution images. Our preferred model, $M_6$ consistently runs at milli-seconds interval compared to multiple seconds by the full-scale baselines. It is also evident that newer generation processor present in Asus mobile helps in faster execution compared to the lower-end models of Mi and Motorola. 
We also profiled the execution times of the models on CPU of a regular commodity laptop with Intel i5 processor and 8GB RAM @ 2.2GHz without any GPU acceleration. It is encouraging to see that even without GPU, model $M_6$ is able to inpaint approximately 3.3 second compared to 0.9, 1.25, and 0.7 second by \cite{iizuka2017globally, gip, shiftnet} respectively. 
Another encouraging observation is that sub-pixel convolution based upsampling (models $M_1$ and $M_2$) is slower on resource constrained mobile platform than proposed bilinear upsampling followed by efficient convolution. This is attributed to the computationally heavy pixel-shuffle operation in $M_1$ and $M_2$. However, on a more resourceful platform such as CPU, this difference is nullified. This observation further strengthens the pragmatism of using bilinear upsampling based efficient upsampling instead of pixel-shuffle based upsampling. \\
\begin{table}[!t]
\centering
\scriptsize
\caption{\scriptsize Comparing computational requirements of various efficient models derived from the full-scale baseline of GLCIC \cite{iizuka2017globally} for inpainting. Cheaper variants using MobileNet, ShuffleNet and ShuffleNetV2 modules are indexed by $`MobNet'$, $`ShNet'$ and $`ShNetV2'$ superscripts.}
\label{tab_compare_shuffle_mobile}
\begin{tabular}{lccccc}\hline\hline
Method & FLOPs & Params & Mobile & CPU & Memory \\
& (10$^9$) & (10$^6$) & (s) & (s) & (MB)\\\hline
GLCIC & 65.0 & 6.02 & 5.8 & 2.1& 40.1\\ \hline
GLCIC$^{MobNet}$ & 9.8 & 0.68 & 0.52 & 1.1& 4.3 \\
GLCIC$^{ShNet}$ & 11.4 & 0.79 & 0.86 & 1.3 & 5.7\\
GLCIC$^{ShNetV2}$ & 10.6 & 0.70 & 0.68 & 1.2 & 4.9 \\\hline
GLCIC$^{M_6}$(\textbf{Proposed}) & \textbf{7.4} & \textbf{0.54} & \textbf{0.35} & \textbf{0.3} & \textbf{2.6}\\ \hline
\end{tabular}
\end{table}
\subsubsection{Comparison with MobileNet and ShuffleNet}
We also designed cheaper variants of GLCIC baseline using efficient convolution units from MobileNet \cite{mobilenet} and ShuffleNet \cite{shufflenet} and ShuffleNetV2 \cite{zhang2018shufflenet}. However, as discussed earlier, these frameworks were targeted for classification tasks and lack any efficient designs for dilated convolution and upsampling operations. For example, both ShuffleNet and ShuffleNetV2 units are invalid on layers in which the number of input and output channels are not same. This is a common design for any upsampling layer. We could have used usual full-scale dilated and transposed convolution for these three frameworks, but for fair comparison with our compressed networks, we add two modifications to these competing frameworks. Firstly, for dilated convolution, we initially perform a dilated 3$\times$3 depthwise convolution followed by 1$\times$1 pointwise convolution. This, itself can be seen as a novel cheaper way of designing dilated convolution layer. Next, for upsampling, we perform bilinear upsampling followed by separable convolution. With these modified settings, we did not find any marked difference of visual quality between the cheaper models and baselines (samples provided in supplementary material for space constraints). From Table \ref{tab_compare_shuffle_mobile} we see that our recommended model, $M_6$ is much more computationally efficient than MobileNet and ShuffleNet variants and, more importantly, $M_6$ has all the necessary components to be seamlessly used in `image-to-image' translation tasks.
\begin{table*}[!t]
\centering
\scriptsize
\caption{\scriptsize Comparison of PSNR and SSIM for AWGN denoising on BSD68 dataset by full capacity baseline  of DnCNN and our proposed cheaper variant DnCNN$^{M_6}$.}
\label{bsd68_result}
\begin{tabular}{|l|c|c|c|c|c|c|c|c|c|}
\hline
 & Noise Level ($\sigma$) $\rightarrow$ & \multicolumn{2}{c|}{10} & \multicolumn{2}{c|}{15} & \multicolumn{2}{c|}{25} & \multicolumn{2}{c|}{50} \\ \hline
 & \multicolumn{1}{l|}{} & \multicolumn{1}{l|}{PSNR(dB)} & \multicolumn{1}{l|}{SSIM} & \multicolumn{1}{l|}{PSNR(dB)} & \multicolumn{1}{l|}{SSIM} & \multicolumn{1}{l|}{PSNR(dB)} & \multicolumn{1}{l|}{SSIM} & \multicolumn{1}{l|}{PSNR(dB)} & \multicolumn{1}{l|}{SSIM} \\ \hline
\multirow{2}{*}{Methods} & DnCNN & 33.78 & 0.92 & 31.75 & 0.89 & 29.23 & 0.83 & 26.29 & 0.72 \\ \cline{2-10} 
 & DnCNN$^{M_6}$ (\textbf{Proposed}) & 33.66 & 0.92 & 31.50 & 0.88 & 29.11 & 0.82 & 26.10 & 0.71 \\ \hline
\end{tabular}
\end{table*}
\subsection{Image Denoising}
\label{sec_denoising}
In this section we show the applicability of our modules to reduce the computational costs of recent state-of-the-art image denoising networks. Henceforth in all experiments we will be using the design strategy and components from our variant, $M_6$, to realize a cheaper version of a given baseline. We initially experimented with \textit{`DnCNN'} framework of Zhang \textit{et al.} \cite{zhang2017beyond} for synthetic All White Gaussian Noise (AWGN) removal. We term our proposed smaller variant as DnCNN$^{M_6}$.
We also experimented to compress the more recent model of CBDNet \cite{guo2019toward} which showed appreciable performance on real-world unknown noise removal and has immediate applications in today's AI-enable cameras. We term the smaller model as CBDNet$^{M_6}$.\\
\subsubsection{Datasets and Training Details}~\\
\textbf{Synthetic Dataset:} We initially compared the performance of our cheaper realization of DnCNN on synthetic AWGN on the widely used BSD68 \cite{roth2009fields} dataset consisting of 68 test images. We experimented on four different noise levels of $\sigma =10, 15, 25, 50$ and zero mean. We followed DnCNN to use 400 images with size $180 \times 180$ for training the network. Random patches of $40 \times 40$ were sampled for training.  \\
\textbf{Real World Dataset:} We also experimented with datasets perturbed with noise from real life unknown noise distributions usually encountered while capturing pictures with contemporary cameras.  For this, we followed the procedures in CBDNet for training our models. A combination of synthetic noise images and real noisy images (120 from  RENOIR\cite{anaya2018renoir}, 400 images from BSD500\cite{martin2003learning}, 1600 images from Waterloo\cite{ma2016waterloo}, and 1600 images from MIT-Adobe FIve\cite{bychkovsky2011learning}) were used for training.
\begin{table}[!t]
\centering
\scriptsize
\caption{\scriptsize Computational requirements of different denoising networks. DnCNN \cite{zhang2017beyond} and CBDNet \cite{cbdnet} are two different full-scale denoising baselines. Cheaper variants using MobileNet, ShuffleNet and ShuffleNetV2 modules are indexed by $`MobNet`$, $`ShNet`$ and $`ShNetV2`$ superscripts. Interesting to note that CBDNet mainly operates on down-sampled feature space unlike DnCNN which operates on full-resolution. So CBDNet baseline has lower FLOPs compared to DnCNN even though the former has more parameters. }
\label{tab_compare_noise_full_scale}
\begin{tabular}{lccccc}\hline\hline
Method & FLOPs & Params & Mobile & CPU & Memory \\
 & (10$^9$) & (10$^6$) & (s) & (s) & (MB)\\\hline
DnCNN & 36.73 & 0.55 & 3.43 & 0.58 & 7.8 \\
CBDNet & 36.09 & 4.34 & 3.30 & 0.49& 29.4 \\\hline
DnCNN$^{MobNet}$ & 5.73  & 0.08 & 0.34 & 0.20   &0.46\\
DnCNN$^{ShNet}$ & 7.44 & 0.18 & 0.41 & 0.24 & 0.6 \\
DnCNN$^{ShNetV2}$ & 7.30 & 0.14 & 0.37 & 0.22 & 0.5 \\\hline
CBDNet$^{MobNet}$ & 6.23 & 0.6 & 0.40 &0.27 & 3.1 \\
CBDNet$^{ShNet}$ & 8.09 & 0.72 & 0.52 & 0.30 & 4.3 \\
CBDNet$^{ShNetV2}$ & 7.60 & 0.69 & 0.48 & 0.28 & 4.1 \\\hline
\begin{tabular}[c]{@{}l@{}}DnCNN$^{M_6}$\\ (\textbf{Proposed})\end{tabular} & 2.97 & 0.04 & 0.16 & 0.11&0.21 \\
\begin{tabular}[c]{@{}l@{}}CBDNet$^{M_6}$\\ (\textbf{Proposed})\end{tabular} & 4.12 & 0.41 & 0.25 & 0.14 &1.93\\ \hline
\end{tabular}
\end{table}
\subsubsection{Denoising Performance}
\begin{figure*}[!t]
    \centering
    \includegraphics[scale=0.25]{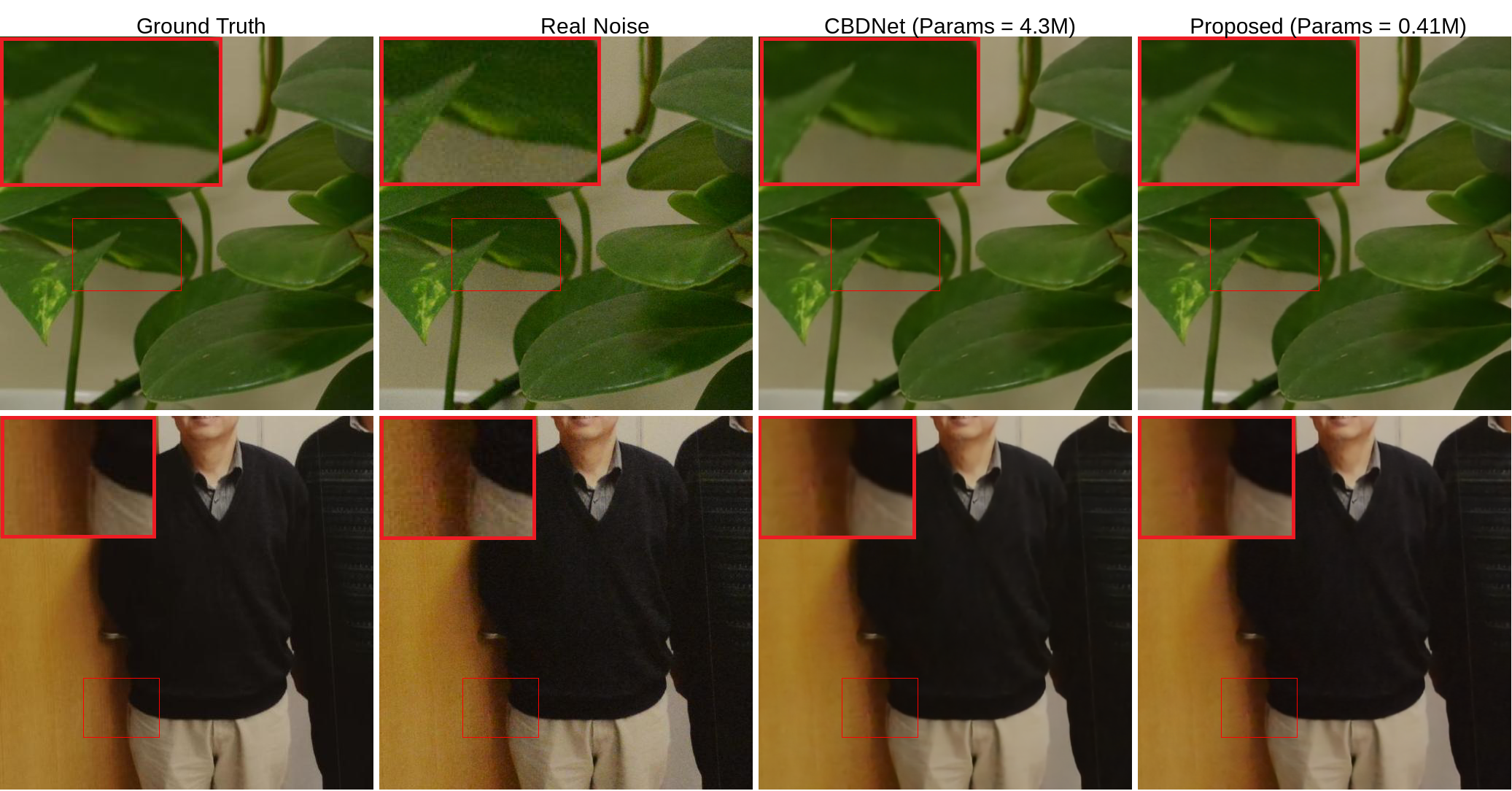}
    \caption{\scriptsize Image denoising on real-world PolyU dataset by full-scale baseline of CBDNet and proposed cheaper variant, CBDNet$^{M_6}$. More examples provided in supplementary material.}
    \label{fig_combined_denoise}
\end{figure*}

~\\
\label{sec_why_mse_report_denoising}
Since all the models are trained to minimize reconstruction loss (instead of adversarial loss), it is pragmatic to compare the models directly in terms of PSNR and SSIM (Structural Similarity Index $\in {0, 1}$) instead of FID. Also, FID calculation requires at least a few thousand samples. However, our test set has a few hundred samples and thus FID metric would not have been a faithful representation of performance.\\
\textbf{Denoising on Synthetic Dataset:}
In Table \ref{bsd68_result} we report the denoising performances of baseline DnCNN and our proposed DnCNN$^{M_6}$ in terms of PSNR and SSIM for AWGN noise removal. SSIM $\in {0, 1}$ is the acronym for Structural Similarity Index. It is used a metric for comparing similarity between two images. SSIM = 1 means perfect match between two images. Across all noise levels, our model has comparable performance to that of DnCNN baseline. \\
\textbf{Denoising on Real Dataset}
For quantitative evaluation we used the publicly available PolyU dataset \cite{xu2018real} containing pairs of real-world noisy and ground truth images. The average PSNR and SSIM for full-scale CBDNet net is 37.95dB and 0.951 while for proposed CBDNet$^{M_6}$ is 37.29dB and 0.948 Again, the differences are not significant. It is encouraging to see that even on real-world noise removal, our compressed variant performs at par with the full-scale CBDNet. Some visual comparisons are provided in Fig. \ref{fig_combined_denoise}.
Additionally, for qualitative evaluations, we used the high-resolution DND \cite{plotz2017benchmarking} dataset in which the ground truths are not publicly available. Due to size limitations we include DND results in \href{https://drive.google.com/open?id=1eO1rKimdpZ17WJcDE_hba1-yP__gtDYx}{this} Google Drive link.\\ 
\textbf{Human Rating:}
In Table \ref{table_mos_denoise} we report the MOS on different datasets. For each dataset, each subject was shown 20 random pairs of noisy and denoised (either from baseline or from our our compressed variant). Total 10 humans participated in the study. The grading strategy (between 0-5) was kept same as that we used during inpainting. We did not find any statistically significant difference (significance set to 10$^{-4}$) between the MOS of baselines and our variant on any of the datasets.\\
\subsubsection{Reduction in Computation}
We report the total number of parameters for the full-scale baseline models of DnCNN and CBDNet and our proposed compressed versions in Table \ref{tab_compare_noise_full_scale}. On DnCNN we achieve 87.27\%  and on CBDNet we achieve 90.2\% relative savings of parameters. Since the models are fully convolutional, any arbitrary resolution of image can be processed. Thus reporting a specific count of FLOPs is not possible. However, for reference, in Table \ref{tab_compare_noise_full_scale} we report the FLOPs for processing input image of resolution 256$\times$256. Proposed CBDNet$^{M_6}$ achieves 89.4\% relative savings in FLOPS compared to CBDNet. We also compare against corresponding compressed variants of DnCNN and CBDNet with MobileNet, ShuffleNet and ShuffleNetV2 modules. Our proposed variant is more efficient in terms of memory requirement and FLOPs compared to both MobileNet and ShuffleNet variants. \\
\subsubsection{Performance on Mobile and CPU:}
In Table \ref{tab_compare_noise_full_scale} we compare the execution times (@ 256$\times$256) on mobile (Asus) and CPU and also the model sizes for mobile deployment. Both of our proposed variants are computationally more economic compared to full-scale baselines as well as MobileNet and ShuffleNet variants.
\par Image denoising is an essential component in majority of contemporary AI-enabled smartphones and the above presented results make our compressed variant a natural substitute for the full-scale models on mobile platforms. 
\begin{table}[!t]
\centering
\scriptsize
\caption{\scriptsize MOS of full-scale baseline of DnCNN \cite{zhang2017beyond} and proposed cheaper variant, DnCNN$^{M_6}$, for AWGN denoising on Set68. For real-world denoising on PolyU dataset we compare baseline of CBDNet \cite{cbdnet} and our cheaper variant, CBDNet$^{M_6}$. Last column shows MOS on original images. }
\label{table_mos_denoise}
\begin{tabular}{l|cccc}\hline \hline
 Dataset & DnCNN & CBDNet & Proposed & Original\\\hline
Set68 ($\sigma = $10) & 4.58  &- & 4.60 & 4.72\\
Set68 ($\sigma = $15) & 4.34  &-  & 4.32 & 4.72\\
Set68 ($\sigma = $25) & 4.10  &- & 4.11 & 4.72\\\hline
Real (PolyU) & - & 4.50 & 4.49 & 4.83\\\hline
\end{tabular}
\end{table}
\subsection{Application in Image Super-Resolution}
\label{sec_sr}
In this section we showcase the efficacy of our modules for single image super-resolution. For this, we consider the benchmark SRGAN model \cite{srgan} as the baseline for 4$\times$ up-scaling. The baseline network consists of series of residual blocks (realized with 3$\times$3 convolution) and upsampling is achieved with sub-pixel convolution with pixel-shuffle operation. We again follow the design principles of our model, $M_6$ to realize cheaper variant of SRGAN.\\ 
\subsubsection{Datasets and Training Details}
We used the training partition of Places2 dataset \cite{places} to train baseline and proposed models. Similar to SRGAN we tested the models on the Set5 \cite{set5}, Set14 \cite{set14}, and BSD 100 (testing set of BSD300 \cite{bsd300}) dataset. Following \cite{srgan}, we randomly cropped 96$\times$96 patch from a given image as HR (high resolution) target and down-sample with bicubic interpolation by 4$\times$ to create the corresponding LR (low resolution) input.
\par We follow the exact same protocol of stagewise training as done in \cite{srgan}. Initially, we train the network with only $\ell_2$ reconstruction loss. The authors term this network as SRResNet. For our smaller model, we term this network as SRResNet$^{M_6}$. Next, we fine-tune the network with VGG-54 content loss and an adversarial loss. Network at this stage is termed at SRGAN for baseline network and SRAGN$^{M_6}$ for our proposed smaller network.\\
\subsubsection{Super-resolution Performance}
\begin{figure*}[!t]
    \centering
    \includegraphics[scale=0.6]{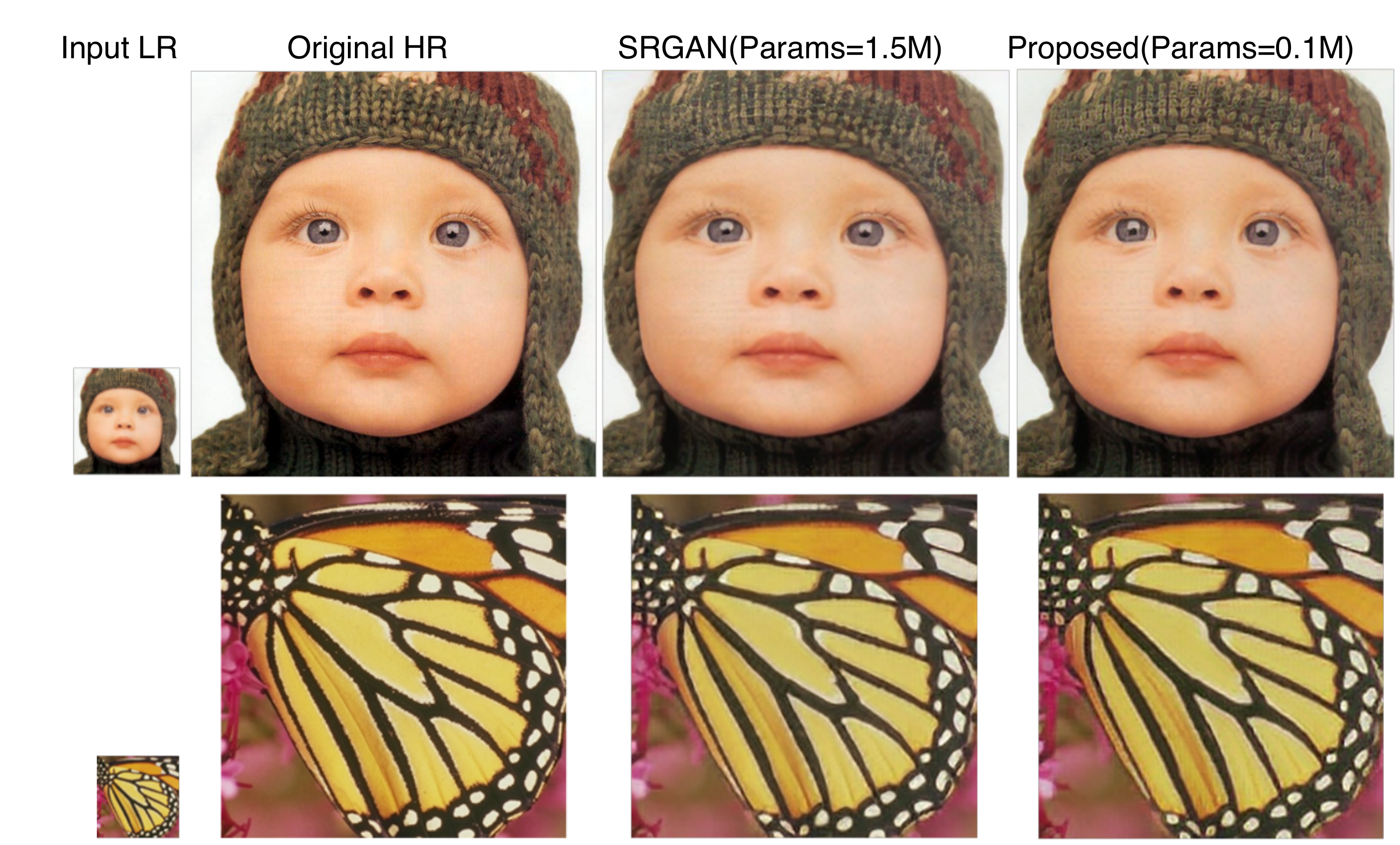}
    \caption{4$\times$ super-resolution by full-scale baseline of $SRGAN$ and our proposed cheaper variant, $SRAGN^{M_6}$.}
    \label{fig_combined_sr}
\end{figure*}

~\\
\textbf{Quantitative Comparison:}
In Table \ref{table_sr_mse} we first compare the PSNR (in dB) of SRResNet and SRResNet$^{M_6}$. Since, both of these models are trained on MSE loss, we compare the PSNR metric. Based on the average PSNR, we could not find any significant difference (significance level set to 10$^{-4}$) between the two models. \\
\textbf{Qualitative Comparison} Next, we conducted a MOS test for the 2 models with 10 independent raters. Each rater was shown the original HR image and the super resolved versions by SRGAN and SRGAN$^{M_6}$  networks. In Table \ref{table_sr_mos} we report the MOS on the three datasets. Again, we could not find any significant difference between the scores received by the models. In Fig. \ref{fig_combined_sr} we visualize some super-resolved images by the two models. It is visually challenging to distinguish samples from the full-scale SRGAN baseline and our cheaper variant. More examples provided in supplementary material. 
\subsubsection{Reduction in Computation}
In Table \ref{table_sr_computational} we report the total number of parameters and FLOPs of different models. FLOPs were calculated on BSD100 dataset in which the original images are usually of dimension 480$\times$320. or 320$\times$480. So, for 4$\times$ super-resolution, input resolution is either 80$\times$120 or 120$\times$80.  Compared to the baseline of SRGAN, our proposed cheaper variant, SRGAN$^{M_6}$ achieves relative parameters and FLOPs savings of 88.4\% and 99\%. Proposed model is also appreciably cheaper compared MobileNet and ShuffleNet variants. 
\begin{table}[!t]
\centering
\scriptsize
\caption{\scriptsize PSNR (in dB) for 4$\times$ super-resolution by baseline SRResNet \cite{srgan} and our proposed cheaper variant SRResNet$^{M_6}$ on Set5, Set14, and BSD100 datasets.}
\label{table_sr_mse}
\begin{tabular}{l|cc}\hline \hline
 Dataset& SRResNet & SRResNet$^{M_6}$\textbf{(Proposed)} \\\hline
Set5 & 31.85 & 31.72 \\
Set14 & 27.90 & 27.74 \\
BSD100 & 27.01 & 26.90\\\hline
\end{tabular}
\end{table}
\begin{table}[!t]
\centering
\scriptsize
\caption{\scriptsize Computational details of different super-resolution networks. SRGAN is the full-scale baseline. Cheaper variants using MobileNet, ShuffleNet and ShuffleNetV2 modules are indexed by $`MobNet`$, $`ShNet`$ and $`ShNetV2`$ superscripts.}
\label{tab_compare_sr_full_scale}
\begin{tabular}{llllll}\hline\hline
Method & FLOPs & Params & Mobile & CPU & Memory \\
& (10$^9$) & (10$^6$) & (s) & (s) & (MB)\\\hline
SRGAN & 38.4 & 1.55 & 3.48 & 0.65 & 12.8 \\
SRGAN$^{MobNet}$ & 0.42 & 0.19 & 0.05 & 0.03  & 1.1\\
SRGAN$^{ShNet}$ & 1.08 & 0.42 & 0.11 & 0.07  & 2.2\\
SRGAN$^{ShNetV2}$ & 1.05 & 0.40 & 0.09 & 0.05  & 1.7\\\hline
\begin{tabular}[c]{@{}l@{}}SRGAN$^{M_6}$\\ \textbf{(Proposed)}\end{tabular} & \textbf{0.27} & \textbf{0.10} & \textbf{0.02} & \textbf{0.01} & \textbf{0.5}\\\hline
\label{table_sr_computational}
\end{tabular}
\end{table}
\begin{table}[!t]
\centering
\scriptsize
\caption{\scriptsize Mean Opinion Score for 4$\times$ super-resolution on outputs of baseline SRGAN-54 \cite{srgan}, our proposed cheaper variant SRGAN-54$^{M_6}$ and original high-resolution images.}
\label{table_sr_mos}
\begin{tabular}{l|ccc}\hline \hline
 Dataset& SRGAN & SRGAN$^{M_6}$ \textbf{(Proposed)} & Original\\\hline
Set5 & 3.62 & 3.64 & 4.45\\
Set14 & 3.69 & 3.72 & 4.41\\
BSD100 & 3.50 & 3.49 & 4.23\\\hline
\end{tabular}
\end{table}
\subsubsection{Performance on Mobiles}
In Table \ref{table_sr_computational} we report the mobile model sizes and execution times on the Asus mobile and the commodity CPU. Proposed variant saves 92.1\%, 47.1\%, 76.1\% and 75.0\% on mobile memory compared to SRGAN-54 baseline, MobileNet, ShuffleNet and ShuffleNetV2 variants respectively. Execution speeds are reported on BSD100 dataset. Our proposed variant achieves significant speedup and reduction of FLOPs compared to full-scale baseline and even MobileNet and ShuffleNet versions.
\section{Conclusion}
In this paper we introduced several convolutional building blocks for low-level restoration tasks. Our proposed modules, \textit{LIST} and \textit{GSAT} were shown to be task agnostic and generalized to variety of restoration tasks. We showed that with specific design consideration, \textit{LIST} layer can be made low cost computationally than contemporary de facto choices of depthwise separable and group convolution based 3$\times$3 layer.
We analytically and empirically analyzed the shortcoming of using depthwise separable kernels to realize sub-pixel convolution based upsampling in an encoder-decoder network configuration. Instead of we showed that homogeneity of network structure can be maintained by deterministic upsampling (instead of transposed convolution or pixel-shuffle based upsampling) followed by efficient convolution with \textit{LIST} layer. Extensive evaluations on resource constrained platforms revealed the effectiveness of our modules in designing computationally efficient yet visually accurate models.
\section*{Acknowledgment}
The work is funded by a Google PhD Fellowship and Qualcomm Innovation Fellowship awarded to Avisek.
\begin{IEEEbiography}[{\includegraphics[width=1in,height=1.25in,clip,keepaspectratio]{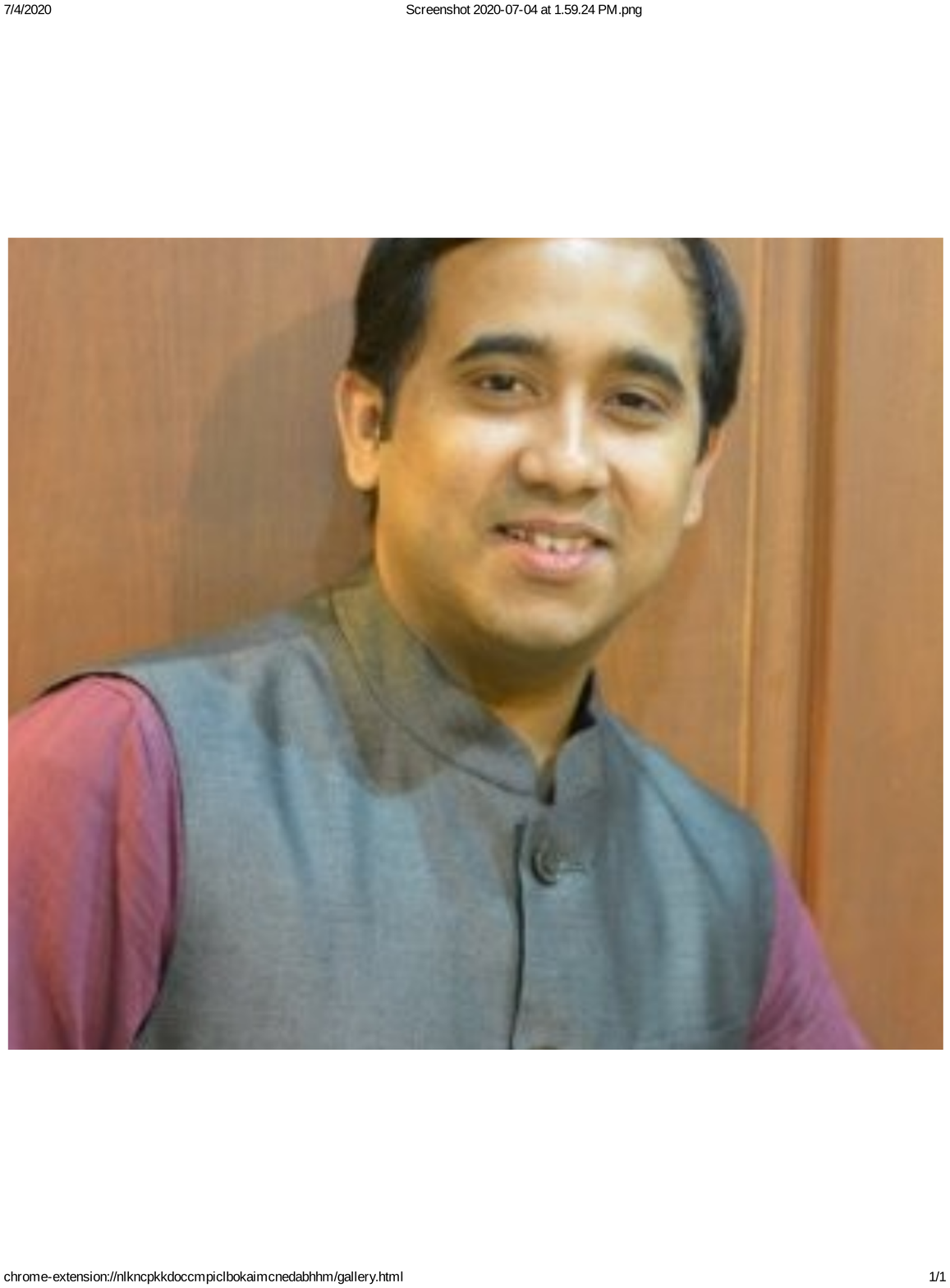}}]%
{Avisek}
is a Ph.D. candidate at the Indian Institute of Technology Kharagpur where he is focusing on image/video reconstruction tasks such as inpainting, super-resolution. His other research interests include data-efficient training of deep neural networks. He is recipient of Google PhD Fellowship and twice recipient of Qualcomm Innovation Fellowship. Avisek was selected as a Young Researcher by the Heidelberg Laureate Forum, 2019. Prior to his Ph.D, Avisek completed his M.S (by research) from IIT Kharagpur with focus on statistical machine learning.
\end{IEEEbiography}
\vskip 0pt plus -1fil
\begin{IEEEbiography}[{\includegraphics[width=1in,height=1.25in,clip,keepaspectratio]{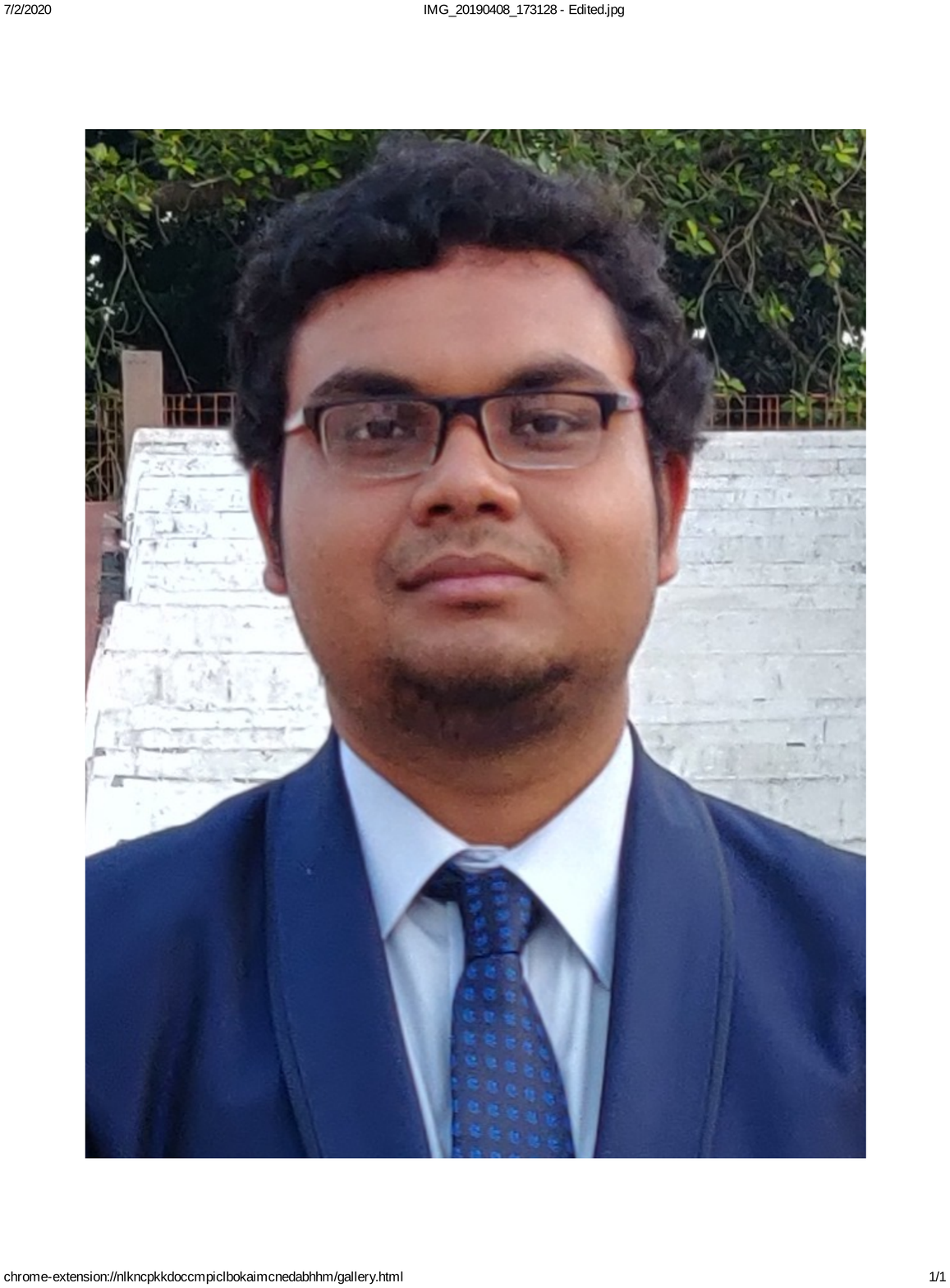}}]%
{Sourav}
 received his M.tech degree from the department of Electronics and Electrical Communication Engineering, IIT Kharagpur, Kharagpur, India.
He is currently working as an advanced deep learning engineer at Mathworks India Pvt. Ltd., Hyderabad. He received the "Institute Silver Medal" for his academic performance at IIT Kharagpur. He received the "Best Student Award" with a gold medal for academic performance during his B.Tech at Kalyani Government Engineering College, Kalyani, India. He has also received many scholarships and awards from the Government of West Bengal for securing 3$^{rd}$ rank at state level in Class X Board examination (Madhyamik) and 8$^{th}$ rank at state level in Class XII Board examination (Higher Secondary).
His current research interest lies in deep learning, machine vision and generative adversarial models.
\end{IEEEbiography}
\vskip 0pt plus -1fil
\begin{IEEEbiography}[{\includegraphics[width=1in,height=1.25in,clip,keepaspectratio]{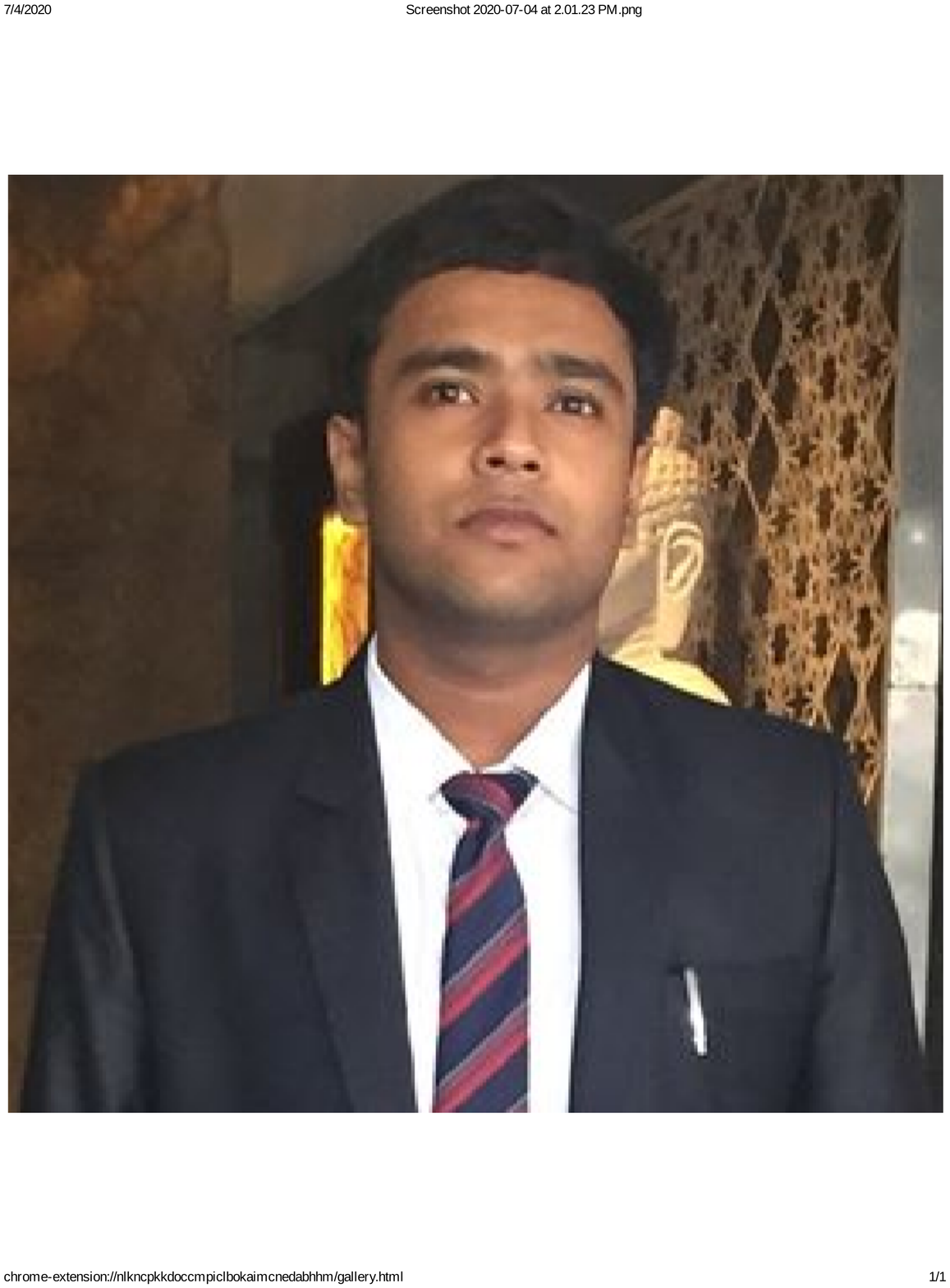}}]%
{Sutanu}
 received BS degree in Electronics and Communication Engineering from the West Bengal University of Technology in 2015 and the MS degree in Medical Imaging and Informatics from the Indian Institute of Technology Kharagpur in 2018. He was awarded the Institute Silver Medal at the time of M.Tech. He is currently a Ph.D. student in the Department of Electrical and Electronics Communication Engineering, Indian Institute of Technology Kharagpur.
His current research interest is deep learning for image restoration, medical image restoration, low-level image processing.
\end{IEEEbiography}
\vskip 0pt plus -1fil
\begin{IEEEbiography}[{\includegraphics[width=1in,height=1.25in,clip,keepaspectratio]{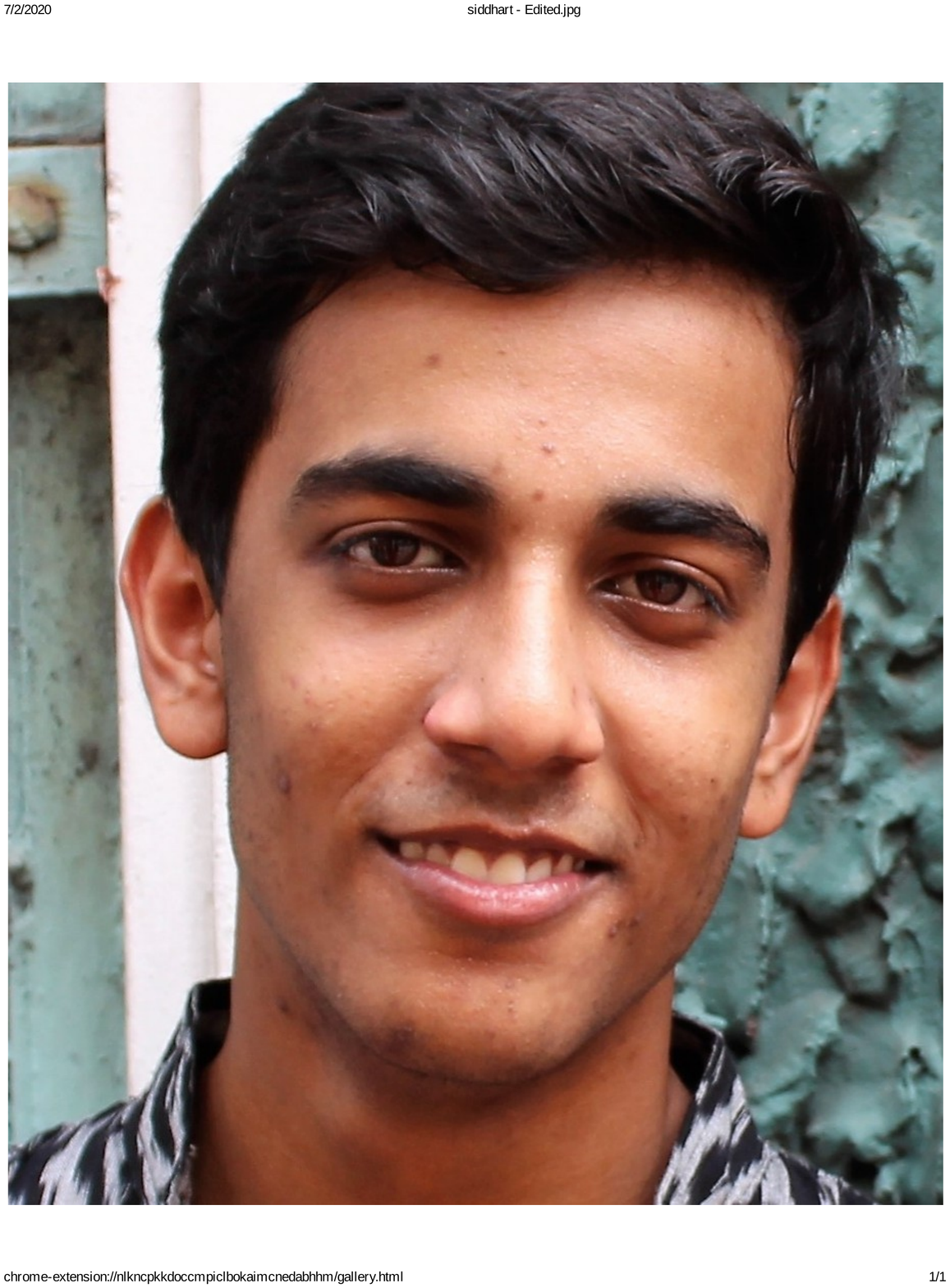}}]%
{Siddhant}
is a final year integrated master's student at Indian Institute of Technology Kharagpur. He is majoring in Electrical Engineering with a minor in Computer Science. He has been doing research in the domain of computer vision, natural language processing, adversarial attacks and causal inference.
\end{IEEEbiography}
\vskip -3pt plus -1fil
\begin{IEEEbiography}[{\includegraphics[width=1in,height=1.25in,clip, keepaspectratio]{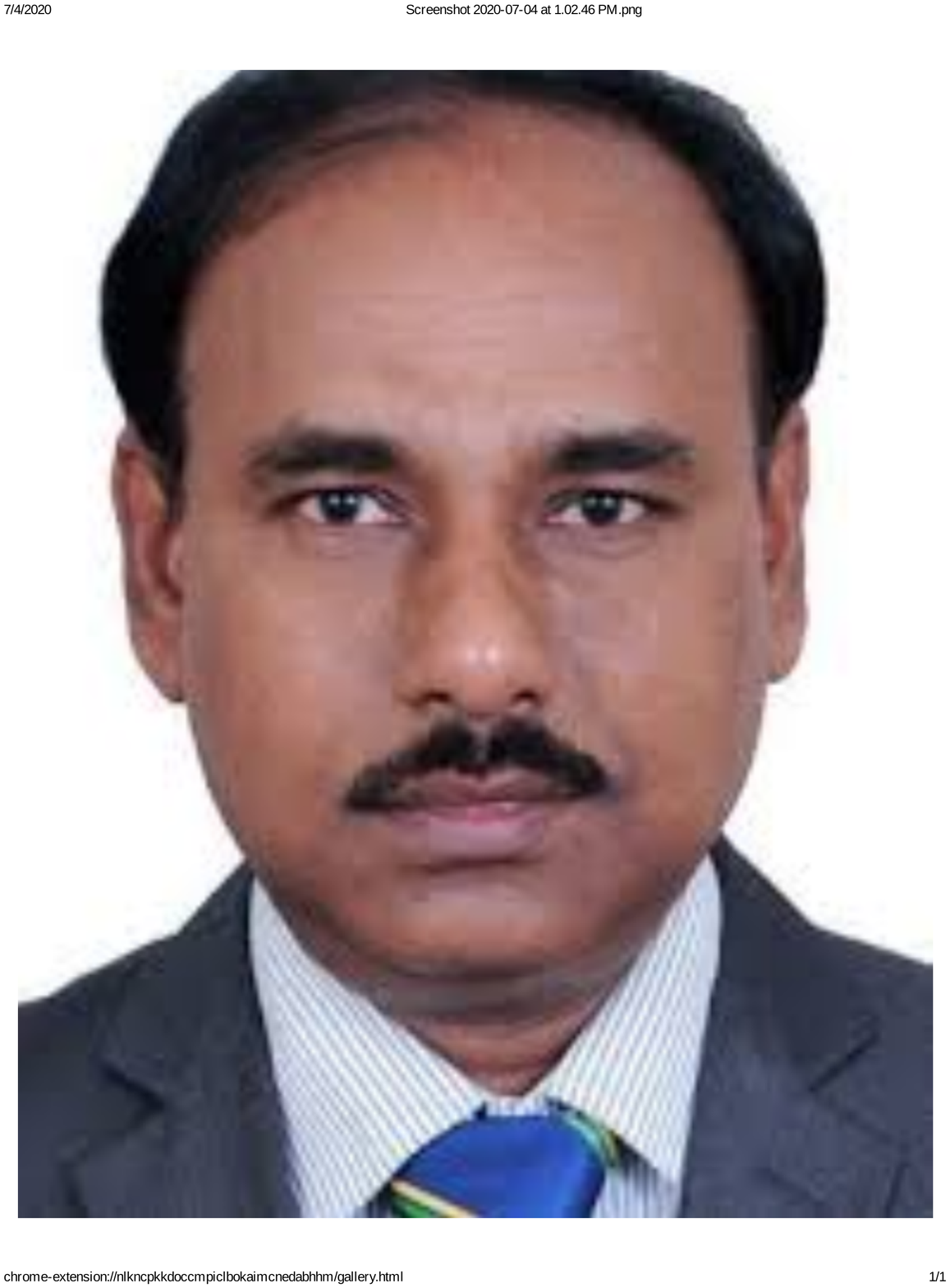}}]%
{Prof. Prabir Kumar Biswas}
(M’93–SM’03) received the B.Tech. (Hons.), M.Tech., and Ph.D. degrees from IIT Kharagpur, Kharagpur, India, in 1985, 1989, and
1991, respectively. 
He was a Visiting Fellow with the University of Kaiserslautern, Kaiserslautern, Germany, under the Alexander von Humboldt Research Fellowship from 2002 to 2003. Since 1991, he has been a Faculty Member with the Department of Electronics and
Electrical Communication Engineering, IIT Kharagpur, where he is currently a Professor, and is the Head of the Department. He has authored over 100 research publications in international and national journals and conferences, and has filed seven international patents. His current research interests include
image processing, pattern recognition, computer vision, video compression, parallel and distributed processing, and computer networks.
\end{IEEEbiography}
{\small
\bibliographystyle{ieee}
\bibliography{egbib}
}
\end{document}